%% file: ms.tex
\documentclass{article}
\pdfoutput=1 

    \PassOptionsToPackage{numbers, compress}{natbib}
    \usepackage[final]{neurips_2019}

\usepackage{amsthm}
\input{defns.tex}

\usepackage{amsmath}

\usepackage[utf8]{inputenc} % allow utf-8 input
\usepackage[T1]{fontenc}    % use 8-bit T1 fonts
\usepackage{hyperref}       % hyperlinks
\usepackage{url}            % simple URL typesetting
\usepackage{booktabs}       % professional-quality tables
\usepackage{amsfonts}       % blackboard math symbols
\usepackage{nicefrac}       % compact symbols for 1/2, etc.
\usepackage{microtype}      % microtypography

\usepackage{color}
\usepackage{todonotes}

\usepackage{thmtools}
\usepackage{thm-restate}
\usepackage{cleveref}

\usepackage{caption}
\usepackage{subcaption}

\input{macros.tex}
\RequirePackage{algorithm}
\RequirePackage{algorithmic}

\title{Approximate Inference Turns Deep Networks into Gaussian Processes}

\makeatletter
\def\@fnsymbol#1{\ensuremath{\ifcase#1\or \dagger\or \dagger\or
   \mathsection\or \mathparagraph\or \|\or **\or \dagger\dagger
   \or \ddagger\ddagger \else\@ctrerr\fi}}

\author{%
  Mohammad Emtiyaz Khan \\
  RIKEN Center for AI Project\\
  Tokyo, Japan\\
  \texttt{emtiyaz.khan@riken.jp} \\
   \And
   Alexander Immer\rlap{\textsuperscript{\normalfont *}}\enskip\thanks{Equal contribution. \textsuperscript{\normalfont *}This work is performed during an internship at the RIKEN Center for AI project.}\\
   EPFL \\
   Lausanne, Switzerland \\
   \texttt{alexander.immer@epfl.ch} \\
   \AND
  Ehsan Abedi\rlap{\textsuperscript{\normalfont *}}\enskip\footnotemark[1] \\
  EPFL\\
   Lausanne, Switzerland \\
   \texttt{ehsan.abedi@epfl.ch} \\
   \And
   Maciej Korzepa\rlap{\textsuperscript{\normalfont *}}\enskip\footnotemark[1] \\
   Technical University of Denmark \\
   Kgs. Lyngby, Denmark \\
   \texttt{mjko@dtu.dk} \\
}

\begin{document}

\maketitle

\begin{abstract}
Deep neural networks (DNN) and Gaussian processes (GP) are two powerful models with several theoretical connections relating them, but the relationship between their training methods is not well understood.
In this paper, we show that certain Gaussian posterior approximations for Bayesian DNNs are equivalent to GP posteriors.
This enables us to relate solutions and iterations of a deep-learning algorithm to GP inference.
As a result, we can obtain a GP kernel and a nonlinear feature map while training a DNN. 
Surprisingly, the resulting kernel is the neural tangent kernel.
We show kernels obtained on real datasets and demonstrate the use of the GP marginal likelihood to tune hyperparameters of DNNs. 
Our work aims to facilitate further research on combining DNNs and GPs in practical settings.
\end{abstract}

\section{Introduction}
\label{sec:introduction}
\input{chapters/_1introduction}

\section{Deep Neural Networks (DNNs) and Gaussian Processes (GPs)}
\label{sec:background}
\input{chapters/_2background}

%\section{Laplace Approximation as GP Inference}
\section{Relating Minima of the Loss to GP Inference via Laplace Approximation}
\label{sec:laplace_dlgp}
\input{chapters/_3laplace_dlgp}

%\section{Variational Inference as GP Inference}
\section{Relating Iterations of a Deep-Learning Algorithm to GP Inference via VI}
\label{sec:vi_dlgp}
\input{chapters/_4vi_dlgp}

\section{Experimental Results}
\label{sec:experiments}
\input{chapters/_5experiments_applications}

\section{Discussion and Future Work}
\label{sec:discussion}
\input{chapters/_6discussion}
%\newpage
\vspace{1em}
\newline
{\bf Acknowledgements}

We would like to thank Kazuki Osawa (Tokyo Institute of Technology), Anirudh Jain (RIKEN), and Runa Eschenhagen (RIKEN) for their help with the experiments. We would also like to thank Matthias Bauer (DeepMind) for discussions and useful feedback. Many thanks to Roman Bachmann (RIKEN) for helping with the visualization in Fig.~\ref{fig:sketch}. We also thank Stephan Mandt (UCI) for suggesting the marginal likelihood experiment. We thank the reviewers and the area chair for their feedback as well. We are also thankful for the RAIDEN computing system and its support team at the RIKEN Center for Advanced Intelligence Project which we used extensively for our experiments.

\bibliographystyle{plainnat}
\bibliography{bibliography.bib}

\clearpage
%\onecolumn
\appendix
{\allowdisplaybreaks
\input{appendix/_appendix}
}

\end{document}

%% file: defns.tex
\newcommand{\loss}{\ell}

\newcommand{\vparam}{\vw}
\newcommand{\param}{w}

\newcommand{\ty}{\tilde{y}}
\newcommand{\tvy}{\tilde{\vy}}
\newcommand{\comment}[1]{}

\newcommand\cut[1]{}

\newcommand{\tvSigma}{\widetilde{\vSigma}}

\newcommand{\squishlist}{
   \begin{list}{$\bullet$}
    { \setlength{\itemsep}{0pt}      \setlength{\parsep}{3pt}
      \setlength{\topsep}{3pt}       \setlength{\partopsep}{0pt}
      \setlength{\leftmargin}{1.5em} \setlength{\labelwidth}{1em}
      \setlength{\labelsep}{0.5em} } }

\newcommand{\squishlisttwo}{
   \begin{list}{$\bullet$}
    { \setlength{\itemsep}{0pt}    \setlength{\parsep}{0pt}
      \setlength{\topsep}{0pt}     \setlength{\partopsep}{0pt}
      \setlength{\leftmargin}{2em} \setlength{\labelwidth}{1.5em}
      \setlength{\labelsep}{0.5em} } }

\newcommand{\squishend}{
    \end{list}  }

{}
{}
{}
{}

{}
{}

{}

\newcommand{\half}{\mbox{$\frac{1}{2}$}}

\newcommand{\real}{\mbox{$\mathbb{R}$}}

\newcommand{\given}{\mbox{$|$}}

\newcommand{\rnd}[1]{\left(#1\right)}
\newcommand{\sqr}[1]{\left[#1\right]}
\newcommand{\crl}[1]{\left\{#1\right\}}

\newcommand{\myexpect}{\mathbb{E}}

\newcommand{\gauss}{\mbox{${\cal N}$}}

\newcommand{\myvec}[1]{\mbox{$\mathbf{#1}$}}
\newcommand{\myvecsym}[1]{\mbox{$\boldsymbol{#1}$}}

\newcommand{\vepsilon}{\mbox{$\myvecsym{\epsilon}$}}
\newcommand{\veta}{\mbox{$\myvecsym{\eta}$}}

\newcommand{\vkappa}{\mbox{$\myvecsym{\kappa}$}}
\newcommand{\vmu}{\mbox{$\myvecsym{\mu}$}}

\newcommand{\vLambda}{\mbox{$\myvecsym{\Lambda}$}}

\newcommand{\vphi}{\mbox{$\myvecsym{\phi}$}}

\newcommand{\vSigma}{\mbox{$\myvecsym{\Sigma}$}}

\newcommand{\vf}{\mbox{$\myvec{f}$}}
\newcommand{\vg}{\mbox{$\myvec{g}$}}
\newcommand{\vh}{\mbox{$\myvec{h}$}}

\newcommand{\vm}{\mbox{$\myvec{m}$}}

\newcommand{\vr}{\mbox{$\myvec{r}$}}
\newcommand{\vs}{\mbox{$\myvec{s}$}}

\newcommand{\vv}{\mbox{$\myvec{v}$}}
\newcommand{\vw}{\mbox{$\myvec{w}$}}

\newcommand{\vx}{\mbox{$\myvec{x}$}}

\newcommand{\vy}{\mbox{$\myvec{y}$}}

\newcommand{\vD}{\mbox{$\myvec{D}$}}

\newcommand{\vH}{\mbox{$\myvec{H}$}}
\newcommand{\vI}{\mbox{$\myvec{I}$}}
\newcommand{\vJ}{\mbox{$\myvec{J}$}}

\newcommand{\vS}{\mbox{$\myvec{S}$}}

\newcommand{\vV}{\mbox{$\myvec{V}$}}

\newcommand{\calD}{\mbox{${\cal D}$}}

\newcommand{\data}{\calD}

\newcommand{\sigmoid}{\mbox{$\sigma$}}

\newcommand{\be}{\begin{equation}}
\newcommand{\ee}{\end{equation}}
\newcommand{\bea}{\begin{eqnarray}}
\newcommand{\eea}{\end{eqnarray}}
\newcommand{\beaa}{\begin{eqnarray*}}
\newcommand{\eeaa}{\end{eqnarray*}}

%% file: chapters/_1introduction.tex
Deep neural networks (DNN) and Gaussian processes (GP) models are both powerful models with complementary strengths and weaknesses. 
DNNs achieve state-of-the-art results on many real-world problems providing scalable end-to-end learning, but they can overfit on small datasets and be overconfident. In contrast, GPs are suitable for small datasets and compute confidence estimates, but they are not scalable and choosing a good kernel in practice is challenging \citep{bradshaw2017adversarial}.
Combining their strengths to solve real-world problems is an important problem.

Theoretically, the two models are closely related to each other.
Previous work has shown that as the width of a DNN increases to infinity, the DNN converges to a GP \citep{Cho&Saul2009,Matthews2018, LeeBNSPS2018, Neal1996, williams1997computing}.
This relationship is surprising and gives us hope that a practical combination could be possible. 
Unfortunately, it is not clear how one can use such connections in practice, e.g., to perform fast inference in GPs by using training methods of DNNs, or to reduce overfitting in DNNs by using GP inference.
We argue that, to solve such practical problems, we need the relationship not only between the models but also between their training procedures. 
The purpose of this paper is to provide such a theoretical relationship.

\begin{figure}[!t]
    \centering
    \includegraphics[width=0.95\textwidth]{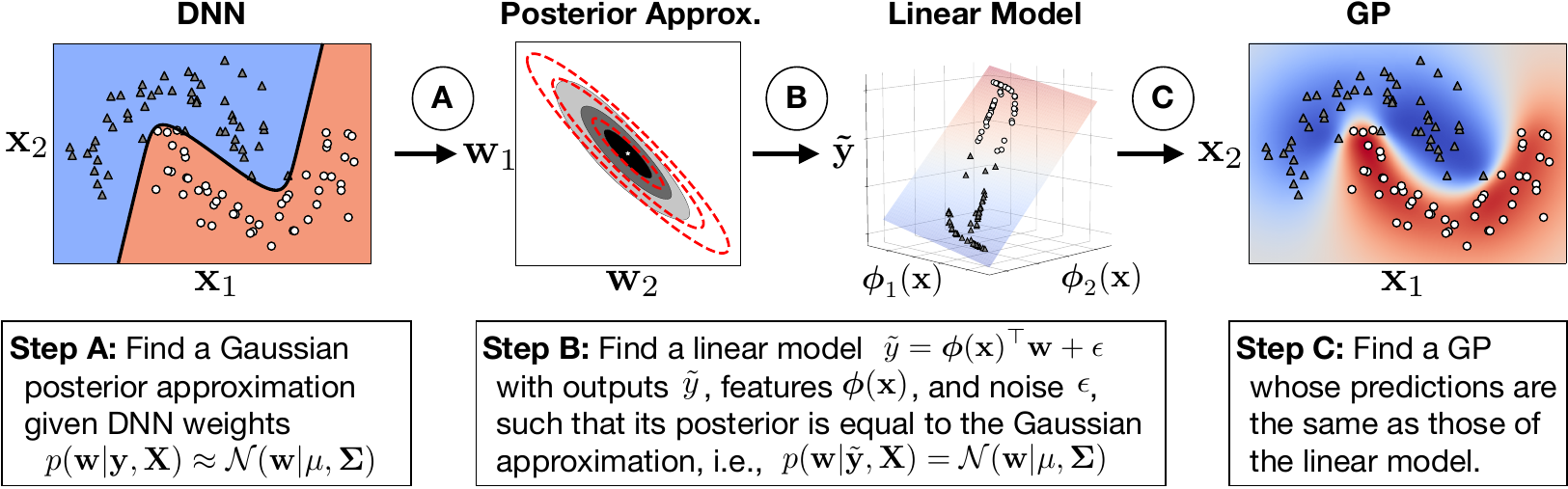}
    \caption{A summary of our approach called DNN2GP in three steps.}
    \label{fig:sketch}
\end{figure}

\begin{figure}
\begin{minipage}[b]{.41\textwidth}
  \vspace*{\fill}
  \centering
  \includegraphics[height=2in]{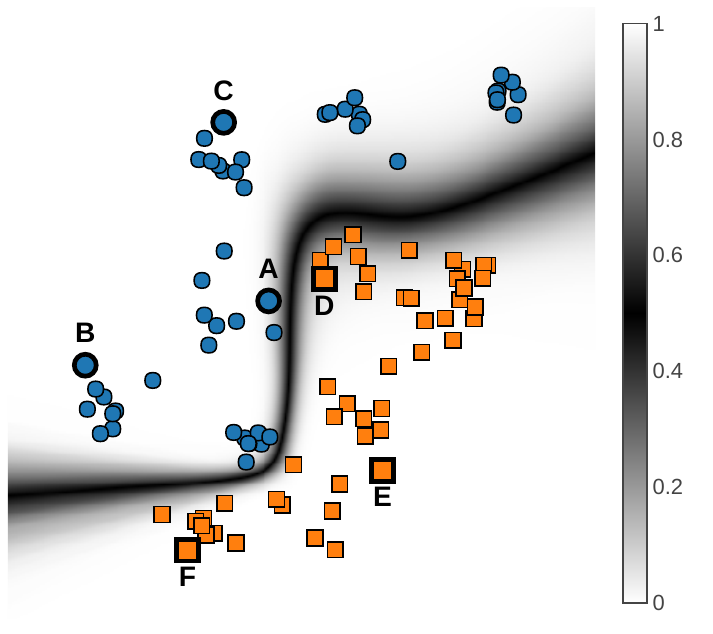}
  \subcaption{2D classification problem}
  \label{fig:test1}
\end{minipage}\hfill
\begin{minipage}[b]{.25\textwidth}
  \vspace*{\fill}
  \centering
  \includegraphics[height=2in]{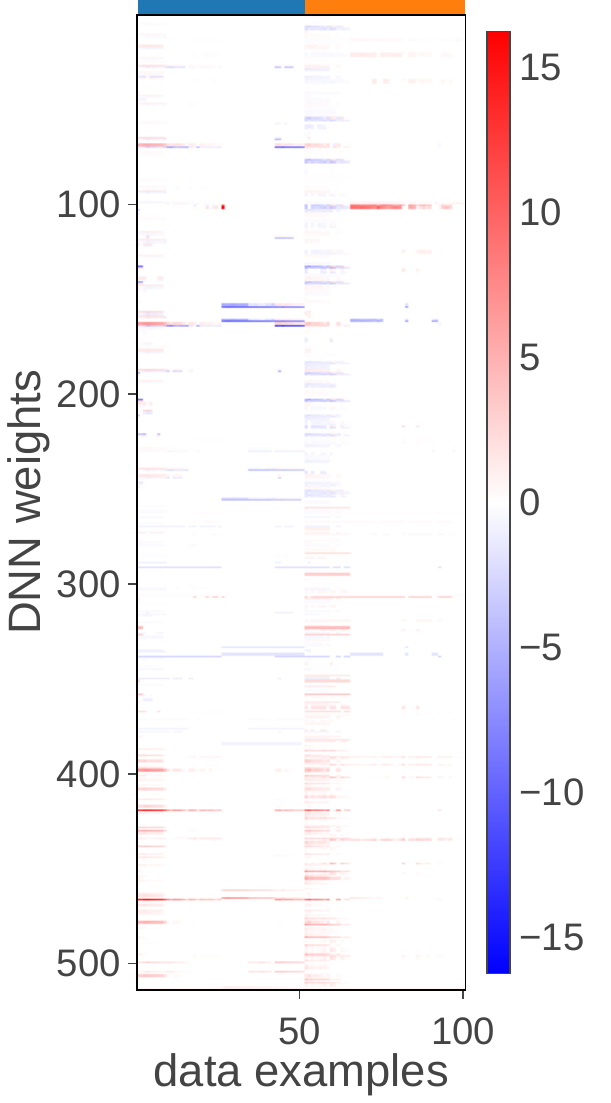}
  \subcaption{GP kernel feature $\vphi(\vx)$}
  \label{fig:test2}
\end{minipage}\hfill
\begin{minipage}[b]{.3\textwidth}
  \vspace*{\fill}
  \centering
  \includegraphics[height=1.46in]{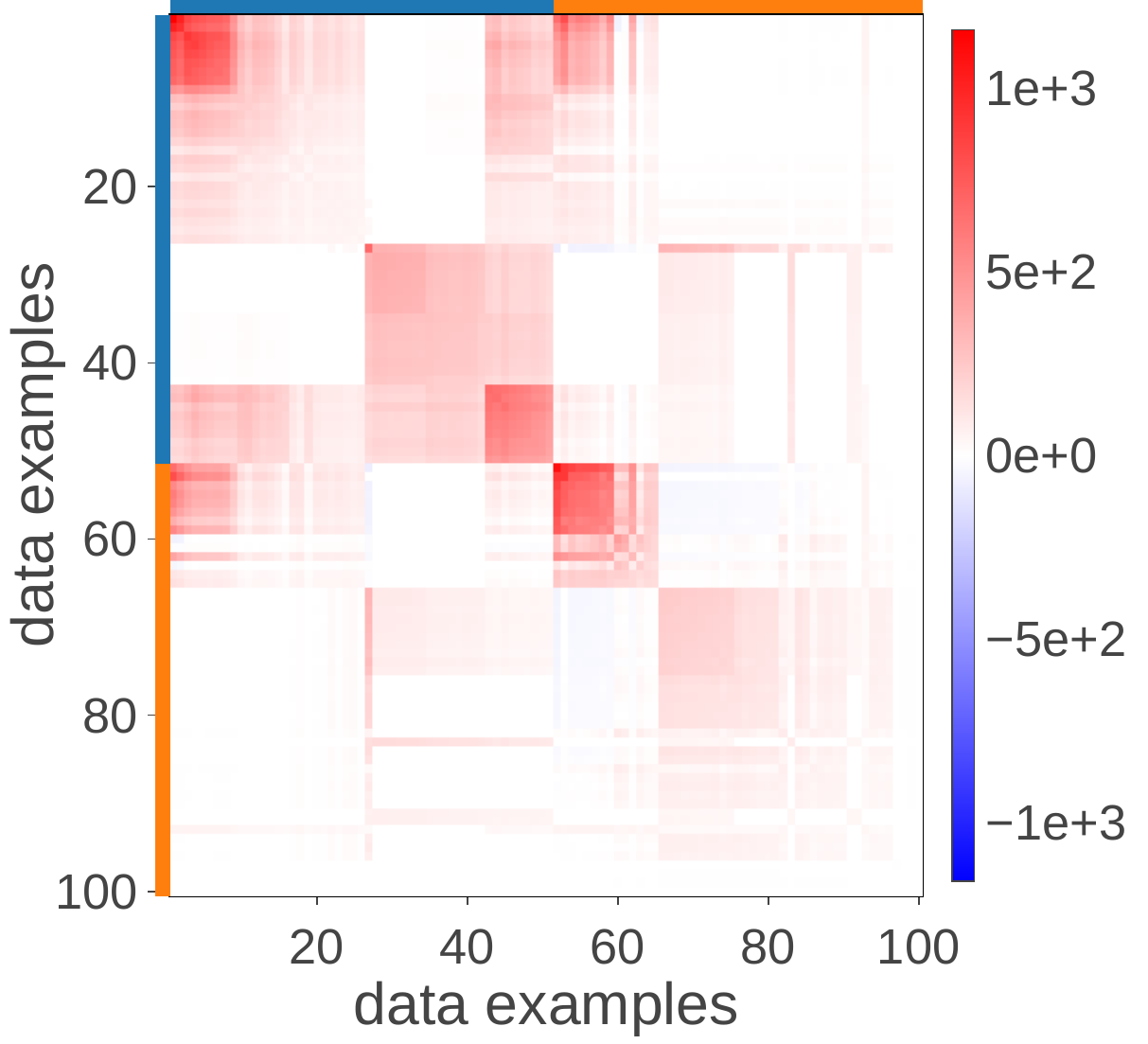}
  \subcaption{GP kernel}
  \label{fig:test3}\par\vfill
  \includegraphics[height=0.3in]{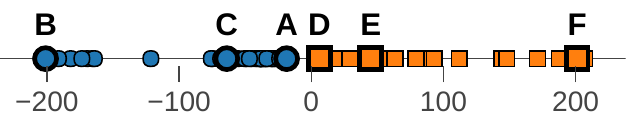}
  \subcaption{GP posterior mean}
  \label{fig:test4}
\end{minipage}
\caption{Fig.~(a) shows a 2D binary-classification problem along with the predictive distribution of a DNN using 513 parameters.
%Our theoretical results show that the approximation is equivalent to the posterior of a GP regression model 
%with kernel $k(\vx_i,\vx_j) := \vJ_*(\vx_i) \vJ_*(\vx_j)^\top$ where $\vJ_*(\vx)^\top\in\real^{513}$ is the Jacobian of the DNN (see Theorem \ref{thm:vi_ggn}). 
The corresponding feature and kernel matrices obtained using our approach are shown in (b) and (c), respectively (the two classes are grouped, and marked with blue and orange color along the axes).
Fig.~(d) shows the GP posterior mean where we see a clear separation between the two classes. Surprisingly, the border points A and D in (a) are also at the boundary in (d).
} 
%\caption{This figure illustrates how variational inference (VI) turns a DNN into a GP. Figure (a) shows 100 two-dimensional inputs $\vx_i$ with two classes, along with the predictions of DNN using 513 parameters.
%We use a variational Gaussian approximation.
%Our theoretical results show that the approximation is equivalent to the posterior of a GP regression model with kernel $k(\vx_i,\vx_j) := \vJ_*(\vx_i) \vJ_*(\vx_j)^\top$ where $\vJ_*(\vx)^\top\in\real^{513}$ is the Jacobian of the DNN (see Theorem \ref{thm:vi_ggn}). 
%Figure (b) shows the feature matrix whose columns are the Jacobian $\vJ_*(\vx_i)$, and Figure (c) shows the GP kernel (classes are grouped and marked with different colors along the axes). In Figure (d), we show the posterior mean of the GP where we see a clear separation between the two classes. Surprisingly, the border points A and D in (a) are also at the boundary in (d).
%} 
\label{fig:2dexample}
\vspace{-1em}
\end{figure}

We present theoretical results aimed at connecting the training methods of deep learning and GP models.
We show that the Gaussian posterior approximations for Bayesian DNNs, such as those obtained by Laplace approximation and variational inference (VI), are equivalent to posterior distributions of GP regression models.
This result enables us to relate the solutions and iterations of a deep-learning algorithm to GP inference. See Fig. \ref{fig:sketch} for our approach called DNN2GP.
In addition, we can obtain GP kernels and nonlinear feature maps while training a DNN (see Fig. \ref{fig:2dexample}).
Surprisingly, a GP kernel we derive is equivalent to the recently proposed neural tangent kernel (NTK) \citep{Jacot2018NeuralTK}.% which has desirable theoretical properties for infinitely-wide DNNs.
We present empirical results where we visualize the feature-map obtained on benchmark datasets such as MNIST and CIFAR, and demonstrate their use for DNN hyperparameter tuning.
The code to reproduce our results is available at \href{https://github.com/team-approx-bayes/dnn2gp}{\texttt{https://github.com/team-approx-bayes/dnn2gp}}.
The work presented in this paper aims to facilitate further research on combining the strengths of DNNs and GPs in practical settings.

%\begin{figure}
%\begin{minipage}[b]{.3\textwidth}
%  \vspace*{\fill}
%  \centering
%  \includegraphics[width=\textwidth]{figures/toy_a.pdf}
%  \subcaption{DNN: preditive probabilities.}
%  \label{fig:test1}
%\end{minipage}\hfill
%%
%\begin{minipage}[b]{.3\textwidth}
%  \vspace*{\fill}
%  \centering
%  \includegraphics[width=\textwidth]{figures/gp_mean_2d_toy.png}
%  \subcaption{DNN2GP: predictive mean.}
%  \label{fig:test1}
%\end{minipage}\hfill
%%
%\begin{minipage}[b]{.3\textwidth}
%  \vspace*{\fill}
%  \centering
%  \includegraphics[width=\textwidth]{figures/gp_var_2d_toy.png}
%  \subcaption{DNN2GP: predictive variance.}
%  \label{fig:test1}
%\end{minipage}
%%
%\caption{Write something.} 
%\label{fig:2dexample}
%\end{figure}

%\todo[size=\tiny]{Regarding Figure 1, state that we use Thm. 2 with 1 sample, $\vparam_*$, and add GP posterior variance}

\subsection{Related Work}
The equivalence between infinitely-wide neural networks and GPs was originally discussed by \citet{Neal1996}. 
Subsequently, many works derived explicit expressions for the GP kernel corresponding to neural networks \cite{Cho&Saul2009, Hazan&Jaakkola2015steps, Neal1996} and their deep variants \cite{Matthews2018, garriga-alonso2019deep, LeeBNSPS2018, novak2019bayesian}.
These works use a prior distribution on weights and derive kernels by averaging over the prior. 
Our work differs from these works in the fact that we use the \emph{posterior} approximations to relate DNNs to GPs.
Unlike these previous results, our results hold for DNNs of finite width.

A GP kernel we derive is equivalent to the recently proposed Neural Tangent Kernel (NTK) \citep{Jacot2018NeuralTK}, which is obtained by using the Jacobian of the DNN outputs.
For randomly initialized trajectories, as the DNN width goes to infinity, the NTK converges in probability to a deterministic kernel and remains asymptotically constant when training with gradient descent.
\citet{Jacot2018NeuralTK} motivate the NTK by using kernel gradient descent. Surprisingly, the NTK appears in our work with an entirely different approach where we consider approximations of the posterior distribution over weights.
%\todo[size=\tiny]{They show that learning by gradient descent is \textit{equivalent} to kernel gradient descent in function space with respect to the NTK. While we show that learning by a natural-gradient variational inference method is \textit{equivalent} to GP regression in function space with respect to a kernel similar to NTK.}
Due to connections to the NTK, we expect similar properties for our kernel.
Our approach additionally shows that we can obtain other types of kernels by using different approximate inference methods.

In a recent work, \citet{Lee2019WideNN} derive the mean and covariance function corresponding to the GP induced by the NTK. Unfortunately, the model does not correspond to inference in a GP model (see Section 2.3.1 in their paper). Our approach does not have this issue and we can express Gaussian posterior approximations on a Bayesian DNN as inference in a GP regression model. 

%% file: chapters/_2background.tex
The goal of this paper is to present a theoretical relationship between training methods of DNNs and GPs. 
DNNs are typically trained by minimizing an empirical loss between the data and the predictions.
For example, in supervised learning with a dataset $\data:=\{(\vx_i,\vy_i)\}_{i=1}^N$ of $N$ examples of input $\vx_i \in \real^D$ and output $\vy_i \in \real^K$, we can minimize a loss of the following form:
\begin{align}
    \bar{\loss}(\data, \vparam) := \sum_{i=1}^N \loss_i(\vparam) + \half \delta \vparam^\top \vparam, \quad \textrm{ where } \loss_i(\vparam) := \loss( \vy_i, \vf_\param(\vx_i) ), \label{eq:dl_loss}
\end{align}
where $\vf_\param (\vx) \in \real^K$ denotes the DNN outputs with weights $\vparam \in \real^P$,  $\loss(\vy,\vf(\vx))$ denotes a loss function between an output $\vy$ and the function $\vf(\vx)$, and $\delta$ is a small $L_2$ regularizer.\footnote{We can assume that $\delta$ is small enough that it does not affect the DNN's generalization.}. We assume the loss function to be twice differentiable and strictly convex in $\vf$ (e.g., squared loss and cross-entropy loss).
An attractive feature of DNNs is that they can be trained using stochastic-gradient (SG) methods~\citep{kingma2014adam}. Such methods scale well to large data settings.

GP models use an entirely different modeling approach which is based on directly modeling the functions rather than the parameters.
For example, for regression problems with scalar outputs $y_i\in\real$, consider the following linear \emph{basis-function} model with a nonlinear feature-map $\vphi(\vx): \real^D\mapsto \real^P$:
        \begin{align}
        y &= \vphi(\vx)^\top\vparam + \epsilon, \,\, \textrm{ with } \epsilon \sim \gauss(0, \sigma^2), \,\, \textrm{ and } \vparam \sim \gauss(0,\delta^{-1}\vI_P), \label{eq:lin_basis}
        \end{align}
where $\vI_P$ is a $P\times P$ identity matrix and $\sigma^2$ is the output noise variance. Defining the function to be $f(\vx) := \vphi(\vx)^\top \vparam$, the predictive distribution $p(f(\vx_*)|\vx_*,\data)$ at a new test input $\vx_*$ is equal to that of the following model directly defined with a GP prior over $f(\vx)$  \citep{williams2006gaussian}:        
\begin{align}
    y = f(\vx) + \epsilon,
    \,\, \textrm{ with }  f(\vx) \sim \mathcal{GP}\rnd{ 0, \kappa(\vx, \vx')},
    \label{eq:gpregmodel}
\end{align}
%\todo[size=\tiny]{why not using the prior $\gauss(\vparam|0,\delta^{-1}\vI_P)$}
where $\kappa(\vx,\vx') := \myexpect[f(\vx)f(\vx')] = \delta^{-1}\vphi(\vx)^\top\vphi(\vx')$
is the \emph{covariance function} or \emph{kernel} of the GP. 
The function-space model is more general in the sense that it can also deal with infinite-dimensional vector feature maps $\vphi(\vx)$, giving us a \emph{nonparametric} model. 
This view has been used to show that as a DNN becomes infinitely wide it tends to a GP, by essentially showing that averaging over $p(\vparam)$ with the feature map induced by a DNN leads to a GP covariance function \citep{Neal1996}.  

An attractive property of the \emph{function-space} formulation as opposed to the \emph{weight-space} formulation, such as \eqref{eq:dl_loss}, is that the posterior distribution has a closed-form expression.
%Therefore, unlike DNNs, once the feature map $\vphi(\vx)$ is fixed then the posterior distribution over $f(\vx)$ is simply a multivariate Gaussian.
Another attractive property is that the posterior is usually unimodal, unlike the loss $\bar{l}(\data,\vparam)$ which is typically nonconvex. Unfortunately, the computation of the posterior takes $O(N^3)$ which is infeasible for large datasets. GPs also require choosing a good kernel \citep{williams2006gaussian}. Unlike DNNs, inference in GPs remains much more difficult.

To summarize, despite the similarities between the two models, their training methods are fundamentally different.
While DNNs employ stochastic optimization, GPs use closed-form updates.
How can we relate these seemingly different training procedures in practical settings, e.g., without assuming infinite-width DNNs?
In this paper, we provide an answer to this question. 
We derive theoretical results that relate the solutions and iterations of deep-learning algorithms to GP inference.
We do so by first finding a Gaussian posterior approximation (Step A in Fig. \ref{fig:sketch}), then use it to find a linear basis-function model (Step B in Fig. \ref{fig:sketch}) and its corresponding GP (Step C in Fig. \ref{fig:sketch}). % by using the equivalence of \eqref{eq:lin_basis} and \eqref{eq:gpregmodel}.
%For Step A, we use Laplace approximations to relate minima of loss functions to GP inference (see Section \ref{sec:laplace_dlgp}).
%To relate iterations of a deep-learning algorithm, we employ natural-gradient variational inference methods \cite{khan2018fast} (see Section \ref{sec:vi_dlgp}).
We start in the next section with our first theoretical result. 
%Our key idea is to use approximate inference on Bayesian DNNs. Approximations such as the Laplace approximation can be obtained by using the solutions found by standard deep learning methods. 
%By showing an equivalence between the DNN posterior approximation and a GP posterior, we can relate solutions and iterations of training algorithms for DNNs to inference in GP. We start with the Laplace approximation in the next section. 

%% file: chapters/_3laplace_dlgp.tex
In this section, we present theoretical results relating minima of a deep-learning loss \eqref{eq:dl_loss} to inference in GP models.
A local minimizer $\vparam_*$ of the loss \eqref{eq:dl_loss} satisfies the following first-order and second-order conditions \cite{nocedal2006numerical}: $\nabla_{\param} \bar{\loss}(\data,\vparam_*) = 0$ and $\nabla_{\param\param}^2 \bar{\loss}(\data, \vparam_*) \succ 0$. Deep-learning optimizers, such as RMSprop and Adam, aim to find such minimizers, and our goal is to relate them to GP inference.  

{\bf Step A (Laplace Approximation):} To do so, we will use an approximate inference method called the Laplace approximation \citep{bishop2006pattern}. The minima of the loss \eqref{eq:dl_loss} corresponds to a mode of the Bayesian model: $p(\data,\vparam) := \prod_{i=1}^N e^{-\loss_i(\mathbf{\param})} p(\vparam)$  with prior distribution $p(\vparam) := \gauss(\vparam|0,\delta^{-1}\vI_P)$,  assuming that the posterior is well-defined.
The posterior distribution $p(\vparam|\data) = p(\data,\vparam)/p(\data)$ is usually computationally intractable and requires computationally-feasible approximation methods.
The Laplace approximation uses the following Gaussian approximation for the posterior: 
\begin{align}
    p(\vparam|\data) \approx \gauss(\vparam|\vmu, \vSigma), \textrm{ where } \vmu = \vparam_* \textrm{ and } \vSigma^{-1} = \sum_{i=1}^N \nabla_{\param\param}^2 \loss_i(\vparam_*) + \delta \vI_P .
    \label{eq:laplace_approx}
\end{align}
This approximation can be directly built using the solutions found by deep-learning optimizers.

{\bf Step B (Linear Model):} The next step is to find a linear basis-function model whose posterior distribution is equal to the Gaussian approximation \eqref{eq:laplace_approx}.
We will now show that this is always possible whenever the gradient and Hessian of the loss\footnote{For notational convenience, we sometime use $\loss(\vparam)$ to denote $\loss(\vy,\vf_w(\vx))$.} can be approximated as follows: 
\begin{align}
\nabla_{\param} \loss(\vparam) \approx \vphi_\param(\vx) \vv_\param(\vx,\vy), \quad \quad  \nabla_{\param\param}^2 \loss(\vparam) \approx  \vphi_\param(\vx) \vD_\param(\vx,\vy) \vphi_\param(\vx)^\top,
\label{eq:grad_hess_form}
\end{align}
%$\nabla_{\param} \loss(\vparam) \approx \vphi(\vx)^\top \vv$ and $\nabla_{\param\param}^2 \loss(\vparam) \approx 
%\vphi(\vx)^\top \vD \vphi(\vx)$,
where $\vphi_w(\vx)$ is a $P \times Q$ feature matrix with $Q$ as a positive integer, $\vv_w(\vx,\vy)$ is a $Q$ length vector, and $\vD_w(\vx,\vy)$ is a $Q\times Q$ symmetric positive-definite matrix.
We will now present results for a specific choice $\vphi_w, \vv_w$, and $\vD_w$.
Our proof trivially generalizes to arbitrary choices of these quantities.
%We will first state our theoretical results assuming that \eqref{eq:grad_hess_form} holds and then discuss specific approximations where such conditions hold.  

%To do so, we employ a further approximation to the Hessian of the loss $\loss_i(\vparam_*)$ by using a \emph{Generalized-Gauss Newton (GGN) approximation} \citep{nocedal2006numerical, martens2014new, Bottou2018Large-Scale}.
For the loss of form \eqref{eq:dl_loss}, the gradient and Hessian take the following form \cite{martens2014new, nocedal2006numerical}:
\begin{align}
    \nabla_{\param} \loss(\vparam) &= \vJ_\param(\vx)^\top \vr_\param(\vx,\vy), \quad
    \nabla_{\param\param}^2 \loss(\vparam) =
    \vJ_\param(\vx)^\top \vLambda_\param(\vx,\vy) \vJ_\param(\vx) +   \vH_f \vr_\param(\vx,\vy), \label{eq:exact_H}
\end{align}
where $\vJ_{\param}(\vx) := \nabla_\param \vf_{\param}(\vx)^\top$ is a $K\times P$ Jacobian matrix, $\vr_w(\vx,\vy) := \nabla_f \loss(\vy, \vf)$ is the \emph{residual} vector evaluated at $\vf := \vf_\param(\vx)$, $\vLambda_w(\vx,\vy) := \nabla_{ff}^2 \loss(\vy, \vf)$, referred to as the \emph{noise precision}, is the $K\times K$ Hessian matrix of the loss evaluated at $\vf := \vf_\param(\vx)$, and $\vH_f := \nabla_{\param\param}^2 \vf_\param(\vx)$.
The similarity between \eqref{eq:grad_hess_form} and \eqref{eq:exact_H} is striking. In fact, if we ignore the second term for the Hessian $\nabla_{\param\param}^2 \loss(\vparam)$ in \eqref{eq:exact_H}, we get the well-known \emph{Generalized Gauss-Newton} (GGN) approximation \citep{martens2014new, nocedal2006numerical}:
%, which is commonly used for nonlinear least-squares problems
%$\nabla_{\param\param}^2 \loss(\vparam) \approx  \vJ_\param(\vx)^\top \vLambda(\vx,y) \vJ_\param(\vx)$.
%The GGN approximation ignores the second term in the Hessian:
\begin{align}
    \nabla_{\param\param}^2 \loss(\vparam) \approx  \vJ_\param(\vx)^\top \vLambda_\param(\vx,\vy) \vJ_\param(\vx).
    \label{eq:ggn}
\end{align}
This gives us one choice for the approximation \eqref{eq:grad_hess_form} where we can set $\vphi_\param(\vx) := \vJ_\param(\vx)^\top, \vv_\param(\vx,\vy) := \vr_\param(\vx,\vy)$, and $\vD_\param(\vx,\vy) := \vLambda_\param(\vx,\vy)$.

We are now ready to present our first theoretical result. Consider a Laplace approximation \eqref{eq:laplace_approx} but with the GGN approximation \eqref{eq:ggn} for the Hessian. We refer to this as \emph{Laplace-GGN}, and denote it by $\gauss(\vparam|\vmu,\widetilde{\vSigma})$ where $\widetilde{\vSigma}$ is the covariance obtained by using the GGN approximation.
We denote the Jacobian, noise-precision, and residual at $\vparam=\vparam_*$ by $\vJ_*(\vx), \vLambda_*(\vx,\vy)$, and $\vr_*(\vx,\vy)$.
%
%We will now show that with these settings we can find a linear model whose posterior distribution is equal to the following modified version of  the Laplace approximation, which we refer to as the \emph{Laplace-GGN} approximation: 
%\begin{align}
%    p(\vparam|\data) \approx \gauss(\vparam|\vmu, \tvSigma), \textrm{ where } \vmu = \vparam_* \textrm{ and } \tvSigma^{-1} = \sum_{i=1}^N \vJ_*(\vx_i)^\top \vLambda_*(\vx_i,\vy_i) \vJ_*(\vx_i) + \delta \vI_P ,
%    \label{eq:laplace_ggn}
%\end{align}
%where $\vJ_*(\vx_i)$ and $\vLambda_*(\vx_i,\vy_i)$ are the Jacobian and noise precision, respectively, evaluated at $\vparam_*$. 
%
%We are now ready to present the theorem which relates the Laplace-GGN approximation to the posterior distribution of a linear model.
We construct a transformed dataset $\widetilde{\data} =\{(\vx_i,\tilde{\vy}_i)\}_{i=1}^N$ where the outputs $\tilde{\vy}_i\in\real^K$ are equal to $\tvy_i := \vJ_*(\vx_i) \vparam_* - \vLambda_*(\vx_i,\vy_i)^{-1}\vr_*(\vx_i,\vy_i)$.
We consider the following linear model for $\widetilde{\data}$:
    \begin{align}
        \tvy &= \vJ_*(\vx) \vparam + \vepsilon, \,\, \textrm{ with } 
        \vepsilon \sim \gauss(0, (\vLambda_*(\vx,\vy))^{-1}) \,\,
        \textrm{ and } \vparam \sim \gauss(0,\delta^{-1} \vI_P). \label{eq:lin_laplace_ggn} 
     \end{align}
The following theorem states our result. 
\begin{restatable}{thm}{thmlaplaceggn}
    \label{thm:laplace_ggn}
%    {\bf ((Laplace approximation as posterior of a linear model): }
        The Laplace approximation $\gauss(\vparam|\vmu,\widetilde{\vSigma})$ is equal to the posterior distribution $p(\vparam|\widetilde{\data})$ of the linear model \eqref{eq:lin_laplace_ggn}.
\end{restatable} 
A proof is given in Appendix \ref{proof:laplaceggn}. The linear model uses $\vJ_*(\vx)$ as the nonlinear feature map, and the noise precision $\vLambda_*(\vx,\vy)$ is obtained using the Hessian of the loss evaluated at $\vf_{\param_*}(\vx)$.
The model is constructed such that its posterior is equal to the Laplace approximation and it exploits the quadratic approximation at $\vparam_*$. 
We now describe the final step relating the linear model to GPs.

% Older theorem
%\begin{restatable}{thm}{thmlaplaceggn}
%    \label{thm:laplace_ggn}
%    {\bf (Laplace approximation as a GP): }
%    The Laplace-GGN approximation \eqref{eq:laplace_ggn} is equivalent to the posterior distribution of a linear basis-function model with observations $\tvy_i \in\real^K$:
%        \begin{align}
%            \tvy_i &= \vJ_*(\vx_i)\vparam + \vepsilon_i, \,\, \textrm{ with } 
%            \vepsilon_i \sim \gauss(0, \vLambda_{i,*}^{-1}) \,\,
%            \textrm{ and } \vparam \sim \gauss(0,\delta^{-1} \vI_P),  \label{eq:lin_laplace_ggn}
%        \end{align}
%    where the observations are defined as $\tvy_i := \vJ_*(\vx_i)\vparam_* - \vLambda_{i,*}^{-1}\vr_{i,*}$ and the feature maps are defined as $\vJ_*(\vx_i)^\top$ with $\vJ_*(\vx_i)$, $\vr_{i,*}$, and $\vLambda_{i,*}$ being the Jacobian, residual, and noise precision evaluated at $\vparam_*$.
%    The predictive distribution of this model is equivalent to that of a GP regression model defined with a \emph{multi-dimensional} $K\times K$ kernel:
%    \begin{align}
%        \tvy_i = \vf(\vx_i) + \vepsilon_i,
%        \quad \textrm{ with } \vf(\vx) \sim \mathcal{GP}\rnd{ 0, \delta^{-1} \vJ_*(\vx) \vJ_*(\vx')^\top} .
%    \end{align}
%\end{restatable} 

{\bf Step C (GP Model):}
    To get a GP model, we use the equivalence between the weight-space view shown in \eqref{eq:lin_basis} and the function-space view shown in \eqref{eq:gpregmodel}.
    With this, we get the following GP regression model whose predictive distribution $p(f(\vx_*)|\vx_*, \widetilde{\data})$ is equal to that of the linear model \eqref{eq:lin_laplace_ggn}: 
    \begin{align}
        \tvy = \vf(\vx) + \vepsilon,
        \quad \textrm{ with } \vf(\vx) \sim \mathcal{GP}\rnd{ 0, \delta^{-1} \vJ_*(\vx) \vJ_*(\vx')^\top} .
        \label{eq:gp_ntk}
    \end{align}
Note that the kernel here is a \emph{multi-dimensional} $K\times K$ kernel.
The steps A, B, and C together convert a DNN defined in the weight-space to a GP defined in the function-space. We refer to this approach as ``DNN2GP''.
%The feature map used in DNN2GP is the Jacobian $\vphi_{\param_*}(\vx) := \vJ_*(\vx)^\top$, defined at a minimum $\vparam_*$ of the loss \eqref{eq:dl_loss}, and the precision matrix $\vLambda_*(\vx,\vy)$ for the noise is the Hessian of the loss. 
%These two quantities, along with the residuals $\vr(\vx,\vy)$, are used to defined the data examples $\tilde{\vy}_i$.

The resulting GP predicts in the space of outputs $\tilde{\vy}$ and therefore results in different predictions than the DNN, but it is connected to it through the Laplace approximation as shown in Theorem~\ref{thm:laplace_ggn}.
In Appendix \ref{app:post_pred}, we describe prediction of the outputs $\vy$ (instead of $\tilde{\vy}$) using this GP.
Note that our approach leads to a heteroscedastic GP which could be beneficial.
Even though our derivation assumes a Gaussian prior and DNN model, the approach holds for other types of priors and models.

{\bf Relationship to NTK:} The GP kernel in \eqref{eq:gp_ntk} is the Neural Tangent Kernel \footnote{The NTK corrsponds to $\delta=1$ which implies a standard normal prior on weights.}
(NTK) \citep{Jacot2018NeuralTK} which has desirable theoretical properties. As the width of the DNN is increasing to infinity, the kernel converges in probability to a deterministic kernel and also remains asymptotically constant during training.
Our kernel is the NTK defined at $\vparam_*$ and is expected to have similar properties. 
It is also likely that, as the DNN width is increased, the Laplace-GGN approximation has similar properties as a GP posterior, and can be potentially used to improve the performance of DNNs.
For example, we can use GPs to tune hyperparameters of DNNs. 
%Such applications would make approximate inference methods desirable for overparametrized DNNs.
The function-space view is also useful to understand relationships between data examples. 
Another advantage of our approach is that we can derive kernels other than the NTK. Any approximation of the form \eqref{eq:grad_hess_form} will always result in a linear model similar to \eqref{eq:lin_laplace_ggn}. 
%$\nabla_{\param} \loss_i(\vparam) \approx \vU(\vx_i)^\top \vv_i$ and $\nabla_{\param\param}^2 \loss_i(\vparam) \approx 
%\vU(\vx_i)^\top \vS_i \vU(\vx_i)$, where $\vU_i$ is a $Q \times P$ matrix, $\vv_i$ is a $Q$ length vector, and $\vS_i$ is a $Q\times Q$ symmetric positive-definite matrix.
%
% REMOVE TO SAVE SPACE
%Other versions where Jacobians are replaced by gradients $\vg_{\param}(\vx,\vy) := \nabla_\param \loss(\vparam)$ are also possible, e.g., by choosing
%$\nabla_{\param} \loss(\vparam) = \vg_{\param}(\vx,\vy)$ and $\nabla_{\param\param}^2 \loss(\vparam) \approx  \vg_{\param}(\vx,\vy) \vg_{\param}(\vx,\vy)^\top$. 
%Such modifications can be useful to reduce computations, but at the cost of increasing the approximation error.

%Another way to derive other types of GP formulation is to change the posterior approximation. We will now show a similar result for variational inference.

{\bf Accuracy of the GGN approximation:} This approximation is accurate when the model $\vf_w(\vx)$ can fit the data well, in which case the residuals $\vr_\param(\vx,\vy)$ are close to zero for all training examples and the second term in \eqref{eq:exact_H} goes to zero \citep{Bottou2018Large-Scale, martens2014new, nocedal2006numerical}.
%Take for instance the least-square loss: $\loss(\vy,\vf_\param(\vx)) = \frac{1}{2\sigma^2} \| \vy - \vf_\param(\vx) \|^2$ with $\sigma^2$ as the noise variance, where the residual and noise precision are simply $\vr_\param(\vx,\vy) = \sigma^{-2} (\vf_\param(\vx) - \vy)$ and $\vLambda_\param(\vx,\vy) := \sigma^{-2} \vI_K$ respectively.
%As the fit becomes more accurate, the residuals go to zero, justifying the approximation (see \citep{nocedal2006numerical, martens2014new, Bottou2018Large-Scale}).
The GGN approximation is a convenient option to derive DNN2GP, but, as it is clear from \eqref{eq:grad_hess_form}, other types of approximations can also be used.

%% file: chapters/_4vi_dlgp.tex
In this section, we present theoretical results relating iterations of an RMSprop-like algorithm to GP inference. The RMSprop algorithm \cite{hintonTieleman} uses the following updates (all operations are element-wise): 
\begin{align}
    \vparam_{t+1} \leftarrow \vparam_t - \alpha_t \rnd{\sqrt{\vs_{t+1}} + \Delta}^{-1}\hat{\vg}(\vparam_t) ,
    \quad\quad \vs_{t+1} \leftarrow (1-\beta_t) \vs_t + \beta_t \rnd{\hat{\vg} (\vparam_t)}^2,
    \label{eq:sg_updates}
\end{align} 
where $t$ is the iteration, $\alpha_t>0$ and $0<\beta_t<1$ are learning rates, $\Delta>0$ is a small scalar, and $\hat{\vg}(\vparam)$ is a stochastic-gradient estimate for $\bar{\loss}(\data, \vparam)$ obtained using minibatches. 
Our goal is to relate the iterates $\vparam_t$ to GP inference using our DNN2GP approach, but this requires a posterior approximation defined at each $\vparam_t$.
We cannot use the Laplace approximation because it is only valid at $\vparam_*$.
We will instead use a version of RMSprop proposed in \cite{khan2018fast} for variational inference (VI), which enables us to construct a GP inference problem at each $\vparam_t$.

{\bf Step A (Variational Inference):} The variational online-Newton (VON) algorithm proposed in \cite{khan2018fast} optimizes the variational objective, but takes an algorithmic form similar to RMSprop (see a detailed discussion in \cite{khan2018fast}). Below, we show a batch version of VON, derived using Eq. (54) in \cite{khan2018fast}:
\begin{align}
    \vmu_{t+1} &\leftarrow  \vmu_t - \beta_t (\vS_{t+1} + \delta\vI_P)^{-1}  \myexpect_{q_t(\param)} \sqr{ \nabla_\param \bar{\loss}(\data, \vparam) }  , \label{eq:VON_1} \\ 
    \vS_{t+1} &\leftarrow (1-\beta_t) \vS_t + \beta_t \sum_{i=1}^N \myexpect_{q_t(\param)} \sqr{ \nabla_{\param\param}^2 \loss_i(\vparam)}   ,
    \label{eq:VON_2}
\end{align}
where $\vS_t$ is a scaling matrix similar to the scaling vector $\vs_t$ in RMSprop, and the Gaussian approximation at iteration $t$ is defined as $q_t(\vparam):=\gauss(\vparam|\vmu_t, \vSigma_t)$ where $\vSigma_t := (\vS_t + \delta\vI_P)^{-1}$.
Since there are no closed-form expressions for the expectations, the Monte Carlo (MC) approximation is used.

{\bf Step B (Linear Model):} As before, we assume the choices for \eqref{eq:grad_hess_form} obtained by using the GGN approximation \eqref{eq:ggn}.
%We refer to the ON variant with the GGN approximation as Online GGN or OGGN.
We consider the variant for VON where the GGN approximation is used for the Hessian and MC approximation is used for the expectations with respect to $q_t(\vparam)$. We call this the Variational Online GGN or VOGGN algorithm. A similar algorithm has recently been used in~\cite{osawa2019} where it shows competitive performance to Adam and SGD.

We now present a theorem relating iterations of VOGGN to linear models.
%A similar result holds for OGGN since it is a particular case of VOGGN with one sample. 
We denote the Gaussian approximation obtained at iteration $t$ by $\Tilde{q}_t(\vparam) := \gauss(\vparam|\vmu_t, \widetilde{\vSigma}_t)$ where $\widetilde{\vSigma}_t$ is used to emphasize the GGN approximation.
We present theoretical results for VOGGN with 1 MC sample which is denoted by $\vparam_t \sim \Tilde{q}_t(\vparam)$.
Our proof in Appendix \ref{proof:oggn} discusses a more general setting with multiple MC samples.
Similarly to the previous section, we first define a transformed dataset: $\widetilde{\data}_t := \{(\vx_i,\tilde{\vy}_{i,t})\}_{i=1}^N$ where 
$\tvy_{i,t} := \vJ_{w_t}(\vx_i)\vparam_t - \vLambda_{w_t}(\vx_i,\vy_i)^{-1}\vr_{w_t}(\vx_i,\vy_i)$,
 and then a linear basis-function model:
    \begin{align}
        \tvy_t &= \vJ_{\param_t}(\vx)\vparam + \vepsilon, \textrm{ with }  \vepsilon \sim \gauss(0, (\beta_t\vLambda_{w_t}(\vx,\vy))^{-1})
        \textrm{ and } \vparam \sim \gauss(\vm_t,\vV_t)
        \label{eq:lin_model_oggn}
    \end{align}
with $\vV_t^{-1} := (1-\beta_t)\widetilde{\vSigma}_t^{-1} + \beta_t \delta \vI_P$ and $\vm_t := (1-\beta_t)\vV_t\widetilde{\vSigma}_t^{-1}\vparam_t$. 
The model is very similar to the one obtained for Laplace approximation, but is now defined using the iterates $\vparam_t$ instead of the minimum $\vparam_*$.
The prior over $\vparam$ is not the standard Gaussian anymore, rather a correlated Gaussian derived from $q_t(\vparam)$. 
The theorem below states the result (a proof is given in Appendix \ref{proof:oggn}).
%The following theorem states the equivalence between the OGGN iterations and the posteriors of the above linear model.
\begin{restatable}{thm}{thmoggn}
    \label{thm:oggn}
%    {\bf ((Laplace approximation as posterior of a linear model): }
        The Gaussian approximation $\gauss(\vparam|\vparam_{t+1},\widetilde{\vSigma}_{t+1})$ at iteration $t+1$ of the VOGGN update is equal to the posterior distribution $p(\vparam|\widetilde{\data}_t)$ of the linear model \eqref{eq:lin_model_oggn}.
\end{restatable} 

{\bf Step C (GP Model):} 
The linear model \eqref{eq:lin_model_oggn} has the same predictive distribution as the GP below:
    \begin{align}
        \tvy_t = \vf_t(\vx) + \vepsilon,
        \quad \textrm{ with } \vf_t(\vx) \sim \mathcal{GP}\rnd{ \vJ_{\param_t}(\vx)\vm_t, \vJ_{\param_t}(\vx) \vV_t \vJ_{\param_t}(\vx')^\top}. 
    \end{align}
The kernel here is similar to the NTK but now there is a covariance term $\vV_t$ which incorporates the effect of the previous $q_t(\vw)$ as a prior.
Our DNN2GP approach shows that one iteration of VOGGN in the weight-space is equivalent to inference in a GP regression model defined in a transformed function-space with respect to a kernel similar to the NTK. 
This can be compared with the results in~\cite{Jacot2018NeuralTK}, where learning by plain gradient descent is shown to be equivalent to kernel gradient descent in function-space.
%One can verify that, at convergence when $\vparam_t \to \vparam_*$, this GP posterior converges to the GP posterior defined with NTK in \eqref{eq:gp_ntk}. 
%
Similarly to the Laplace case, the resulting GP predicts in the space of outputs $\tilde{\vy}_t$, but predictions for $\vy_t$ can be obtained using a method described in Appendix \ref{app:post_pred}.

{\bf A Deep-Learning Optimizer Derived from VOGGN:}
The VON algorithm, even though similar to RMSprop, does not converge to the minimum of the loss. This is because it optimizes the variational objective. 
Fortunately, a slight modification of this algorithm gives us a deep-learning optimizer which is similar to RMSprop but is guaranteed to converge to the minimum of the loss. 
For this, we approximate the expectations in the updates \eqref{eq:VON_1}-\eqref{eq:VON_2} at the mean $\vmu_t$.
This is called the \emph{zeroth-order delta approximation}; see Appendix A.6 in \cite{khan2012variational} for details of this method. 
Using this approximation and denoting the mean $\vmu_t$ by $\vparam_t$, we get the following update:
\begin{align}
    \vparam_{t+1} &\leftarrow  \vparam_t - \beta_t (\hat{\vS}_{t+1} + \delta\vI_P)^{-1}  \nabla_\param \bar{\loss}(\data, \vparam_t),  \quad\,\,\,
    \hat{\vS}_{t+1} \leftarrow (1-\beta_t) \hat{\vS}_t + \beta_t \sum_{i=1}^N \sqr{ \nabla_{\param\param}^2 \loss_i(\vparam_t)}.
    \nonumber %\label{eq:ON_2}
\end{align}
We refer to this as Online GGN or OGGN method.
A fixed point $\vparam_*$ of this iteration is also a minimizer of the loss since we have $\nabla_\param \bar{\loss}(\data, \vparam_*)=0$.
Unlike RMSprop, at each iteration, we still get a Gaussian approximation $\hat{q}_t(\vparam) := \gauss(\vparam|\vparam_t, \hat{\vSigma}_t)$ with $\hat{\vSigma}_t := (\hat{\vS}_t + \delta\vI_P)^{-1}$.
Therefore, the posterior of the linear model from Theorem \eqref{thm:oggn} is equivalent to $\hat{q}_t$ when $\widetilde{\vSigma}_t$ is replaced by $\hat{\vSigma}_t$ (see Appendix \ref{app:oggn_proof}). 
In conclusion, by using VI in our DNN2GP approach, we are able to relate the iterations of a deep-learning optimizer to GP inference.
%This is unlike RMSprop where there is no such interpretation of updates resulting in Gaussian approximation. 
%At convergence, this approximation is equal to the Laplace approximation.
%
%A similar result for VOGGN is discussed in Appendix \ref{} \todo[]{add appendix}, where the GP kernel is specified in a very similar fashion but obtained at samples $\vparam \sim q_t(\vparam)$, instead of the iterates $\vparam_t$.
%The sampling encourages Bayesian averaging and is expected to improves uncertainty estimates.

{\bf Implementation of DNN2GP:} In practice, both VOGGN and OGGN are computationally more expensive than RMSprop because they involve computation of full covariance matrices.
To address this issue, we simply use the diagonal versions of these algorithms discussed in \cite{khan2018fast, osawa2019}. 
Specifically, we use the VOGN and OGN algorithms discussed in~\cite{osawa2019}.
This implies that $\vV_t$ is a diagonal matrix and the GP kernel can be obtained without requiring any computation of large matrices.
Only Jacobian computations are required.
In our experiments, we also resort to computing the kernel over a subset of data instead of the whole data, which further reduces the cost.

%% file: chapters/_5experiments_applications.tex
\begin{figure}[t]
     \centering
     \begin{subfigure}[b]{0.49\textwidth}
         \centering
         \includegraphics[height=1.5in]{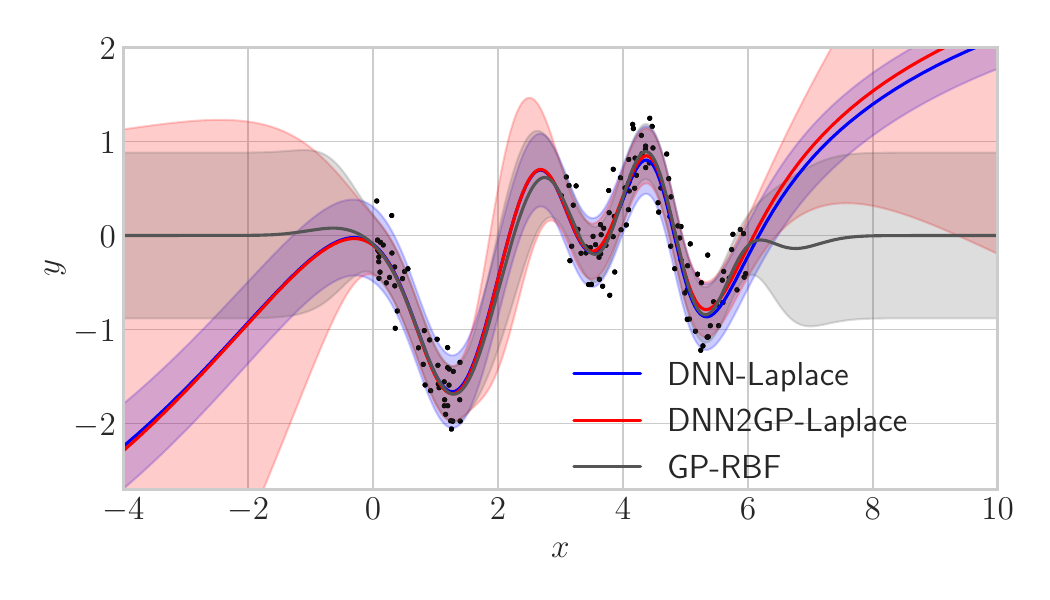}
         \vspace{-9pt}
%         \subcaption{DNN2GP-Laplace}
%         \label{fig:uncertainty_laplace}
     \end{subfigure}
     \hfill
     \begin{subfigure}[b]{0.49\textwidth}
         \centering
         \includegraphics[height=1.5in]{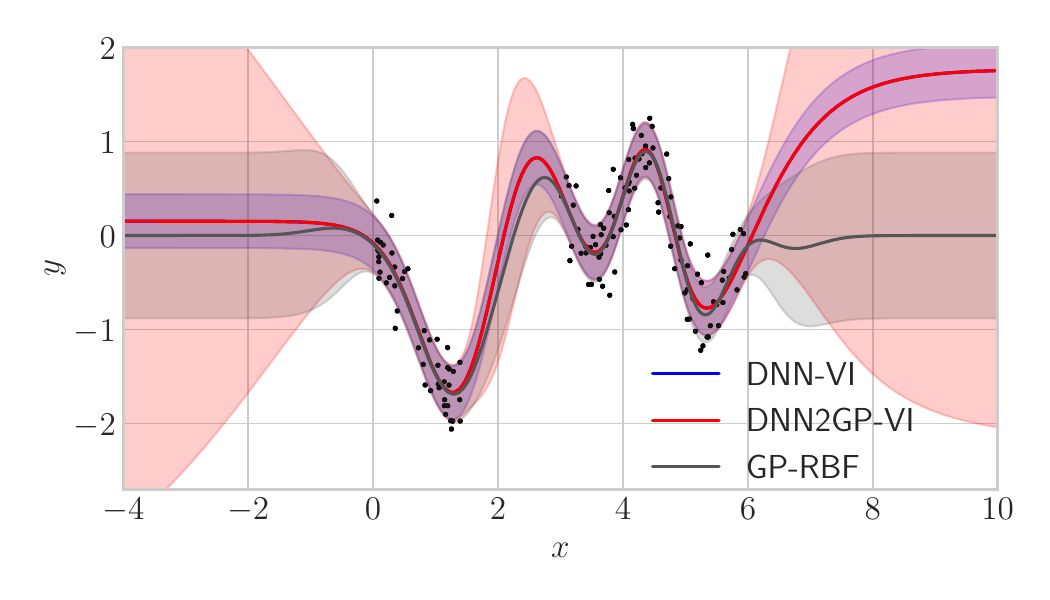}
         \vspace{-9pt}
%         \subcaption{DNN2GP-VI}
%         \label{fig:uncertainty_vi}
     \end{subfigure}
    \vspace{-0.4em}
        \caption{This figure shows a visualization of the predictive distributions on a modified version of the Snelson dataset \cite{Snelson}. 
        The left figure shows Laplace and the right one shows VI.
        DNN2GP is our proposed method, elaborated upon in Appendix~\ref{app:post_pred}, while DNN refers to a diagonal Gaussian approximation. We also compare to a GP with RBF kernel (GP-RBF). An MLP is used for DNN2GP and DNN.
        We see that, wherever the data is missing, the uncertainties are larger for our method than the others.
        For classification, we give an example in Fig.~\ref{fig:bin_uct} in the appendix.
        }
    \label{fig:Snelson}
    \vspace{-1.5em}
\end{figure}

\subsection{Comparison of DNN2GP Uncertainty} \label{eq:exactgp}
In this section, we visualize the quality of the uncertainty of the GP obtained with our DNN2GP approach on a simple regression task. 
To approximate predicitive uncertainty for our approach, we use the method described in Appendix~\ref{app:post_pred}. 
We use both Laplace and VI approximations, referred to as `DNN2GP-Laplace' and `DNN2GP-VI', respectively.
We compare it to the uncertainty obtained using an MC approximation in the DNN (referred to as `DNN-Laplace' and `DNN-VI').
We also compare to a standard GP regression model with an RBF kernel (refer to as `GP-RBF'), whose kernel hyperparameters are chosen by optimizing the GP marginal likelihood. 

We consider a version of the Snelson dataset \cite{Snelson} where, to assess the `in-between' uncertainty, we remove the data points between $x=1.5$ and $x=3$. 
We use a single hidden-layer MLP with 32 units and sigmoidal transfer function.
Fig.~\ref{fig:Snelson} shows the results for Laplace (left) and VI (right) approximation. For Laplace, we use Adam~\citep{kingma2014adam}, and, for VI, we use VOGN~\citep{khan2018fast}.
The uncertainty provided by DNN2GP is bigger than the other methods wherever the data is not observed.
% Several works have shown that as the width of a DNN is increased, it converges to a GP.
% We compare the predictions of our GP model to those of the exact GP corresponding to an infinitely-wide DNN.
% We consider a neural network with a sigmoidal transfer function. 
% An exact expression for the GP is available when the width goes to infinity~\cite{williams1997computing}.
% Using this, we compute predictive distribution of the exact GP.
% We compare it to the posterior predictive distribution of the GP obtained due to Theorem~\ref{thm:laplace_ggn} (see details for the computation of the predictive distribution in Appendix \ref{app:post_pred}).
% We increase the width of the neural network from 2 to 64. 
% Figure \ref{fig:Williams_infinite_width} shows the results on a synthetic dataset, where we see that the two predictive distributions match well for large width.
% \todo[size=\tiny]{In the limit, the predictive distribution by Laplace GP does not converge to the exact GP in theory. There are some papers to compute the NTK of infinite networks exactly. According, we might have a single figure, comparing the exact GP and the Laplace GP in the infinite width limit.}

\subsection{GP Kernel and Predictive Distribution for Classification Datasets} \label{sec:appliedexp}
In this section, we visualize the GP kernel and predictive distribution for DNNs trained on CIFAR-10 and MNIST. Our goal is to show that our GP kernel and its predictions enhance our understanding of a DNN's performance on classification tasks. 
We consider LeNet-5~\citep{lecun1998gradient} and compute both the Laplace and VI approximations. We show the visualization at the posterior mean.

The $K\times K$ GP kernel $\vkappa_*(\vx,\vx') := \vJ_*(\vx)\vJ_*(\vx')^\top$ results in a kernel matrix of dimensionality $NK\times NK$ which makes it difficult to visualize for our datasets.
To simplify, we compute the sum of the diagonal entries of $\vkappa_*(\vx,\vx')$ to get an $N\times N$ matrix. 
This corresponds to modelling the output for each class with an individual GP and then summing the kernels of these GPs.
We also visualize the GP posterior mean: $\myexpect[\vf(\vx)|\data] = \myexpect[\vJ_*(\vx)\vparam|\data] = \vJ_*(\vx)\vparam_* \in \real^K$. 
%If the true label is $k$, then we want the $k$'th entry of the posterior mean to be the highest.
%This is because we expect the GP function $\vf(\vx)$ to be related to the DNN function $\vf_\param(\vx)$ at training data points.
and use the reparameterization that allows to predict in the data space $\vy$ instead of $\Tilde{\vy}$ which is explained in Appendix~\ref{app:post_pred}.

\begin{figure}
\begin{minipage}[b]{.68\textwidth}
  \vspace*{\fill}
  \centering
  \includegraphics[height=1.7in]{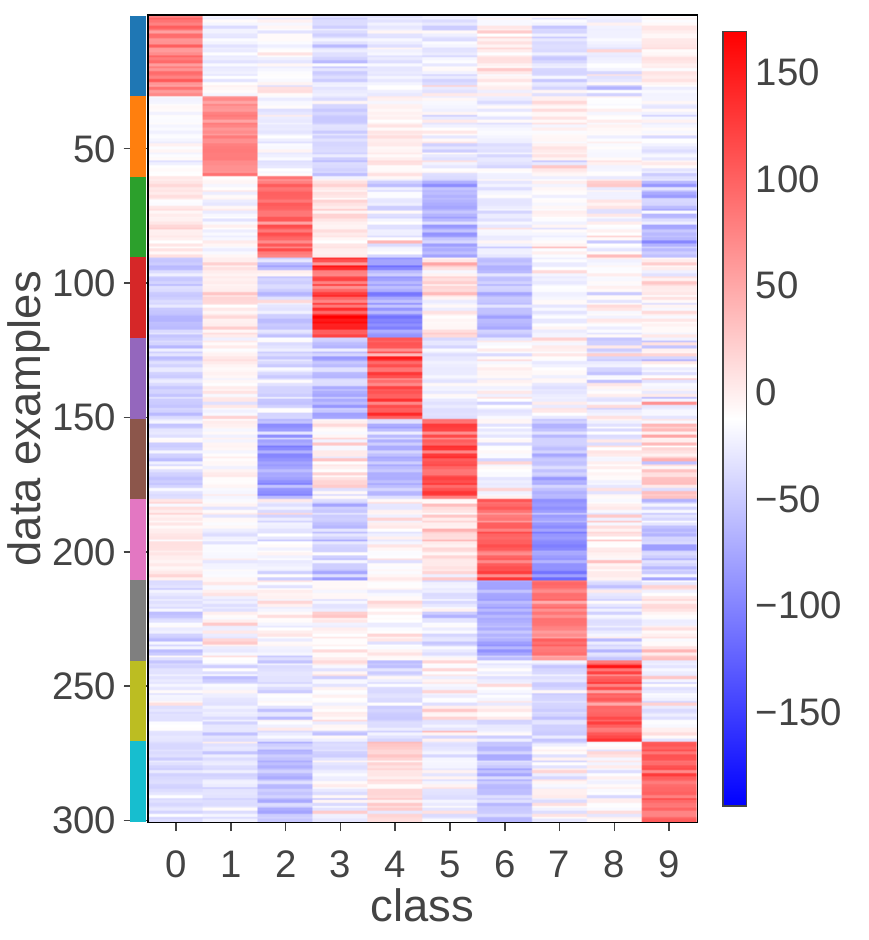}
  \includegraphics[height=1.7in]{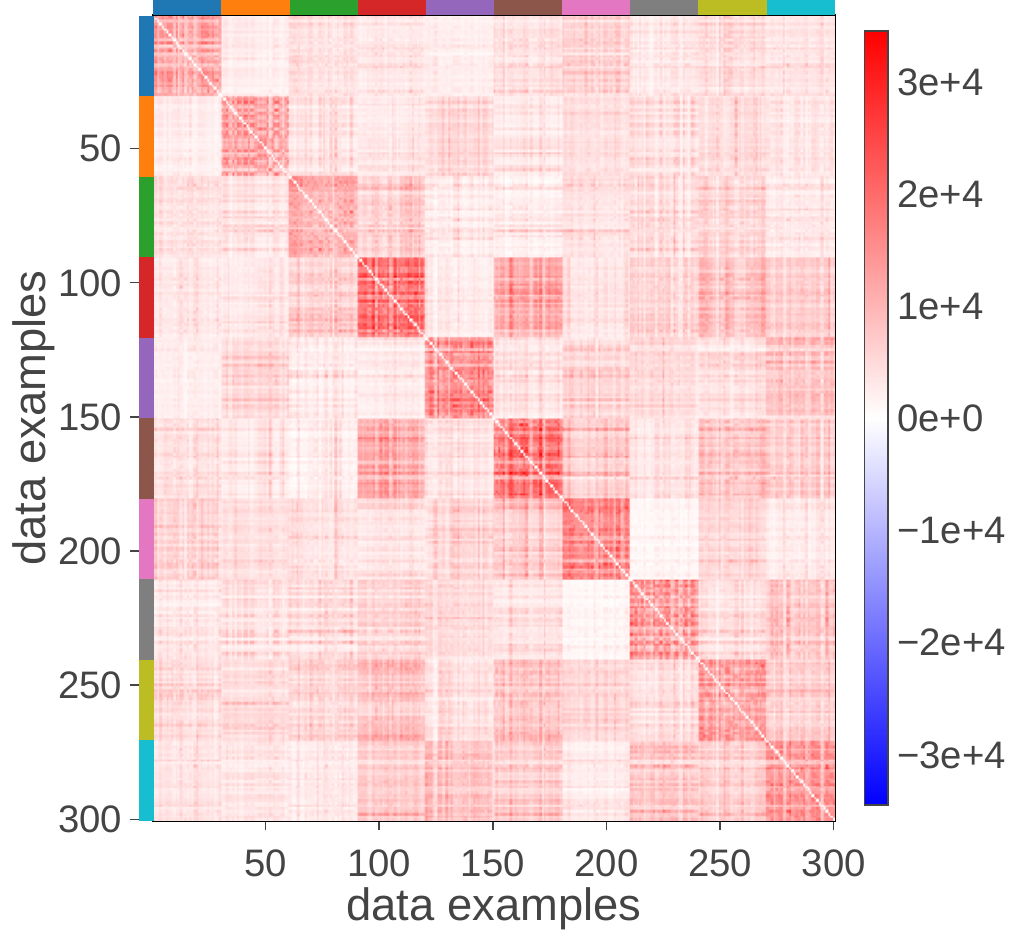}
  \vspace{-3pt}
  \subcaption{MNIST: GP posterior mean (left) and GP kernel matrix (right)}
  \label{fig:mnist_adam}
\end{minipage}\hfill
\begin{minipage}[b]{.3\textwidth}
  \vspace*{\fill}
  \centering
\includegraphics[height=1.7in]{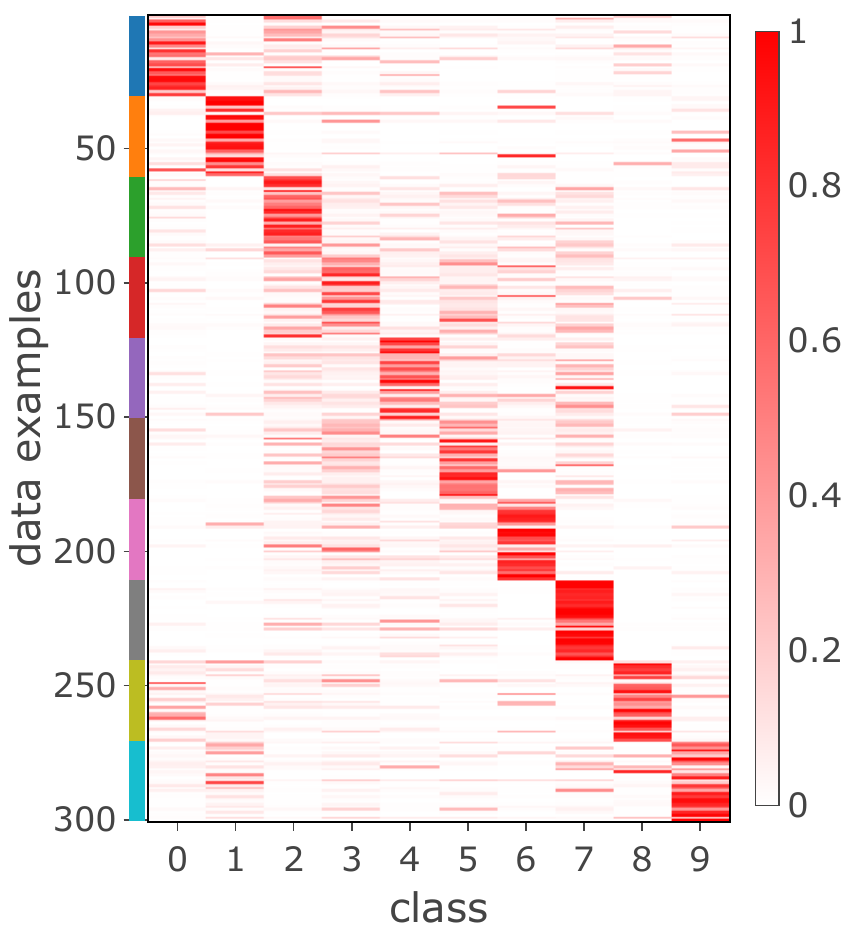}
  \vspace{-3pt}
  \subcaption{CIFAR: GP posterior mean}
  \label{fig:cifar_adam}
\end{minipage} \\
% \vspace{-0.1em}
\begin{minipage}[b]{.49\textwidth}
  \vspace*{\fill}
  \centering
  \includegraphics[height=1.7in]{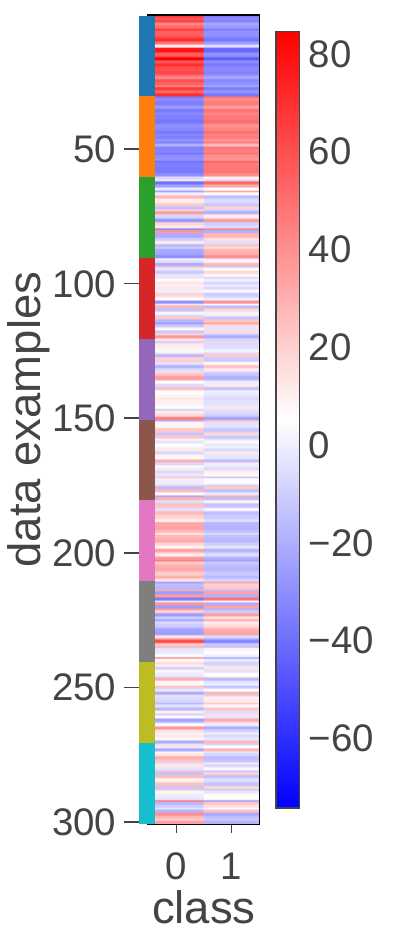}
  \includegraphics[height=1.7in]{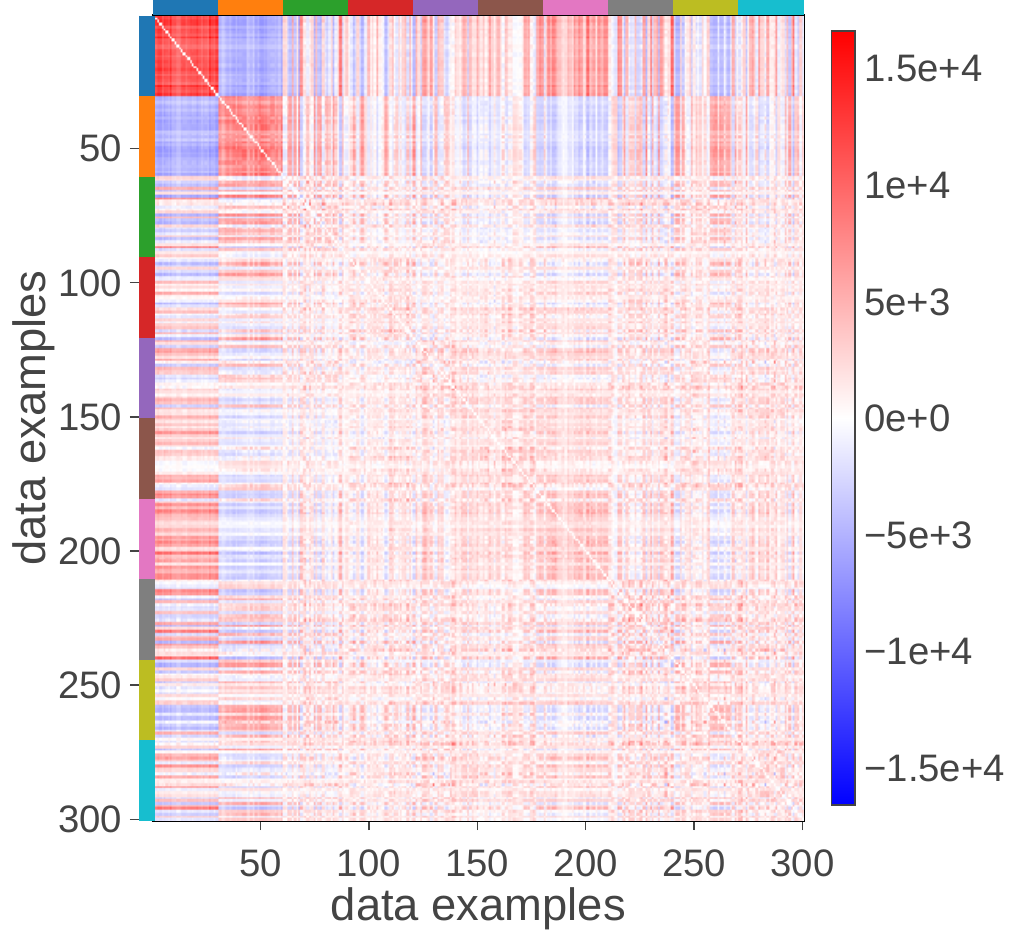}
  \vspace{-3pt}
  \subcaption{Binary-MNIST on digits 0 and 1}
  \label{fig:mnistzone}
\end{minipage}\hfill
% \begin{minipage}[b]{.49\textwidth}
%   \vspace*{\fill}
%   \centering
%   \includegraphics[height=1.7in]{figures/fnine_vogn_obs.pdf}
%   \includegraphics[height=1.7in]{figures/fnine_vogn_ker.pdf}
%   \subcaption{Binary classification on digits 4 and 9}
%   \label{fig:mnistfnine}
% \end{minipage}
\begin{minipage}[b]{.49\textwidth}
   \vspace*{\fill}
   \centering
   \includegraphics[height=1.7in]{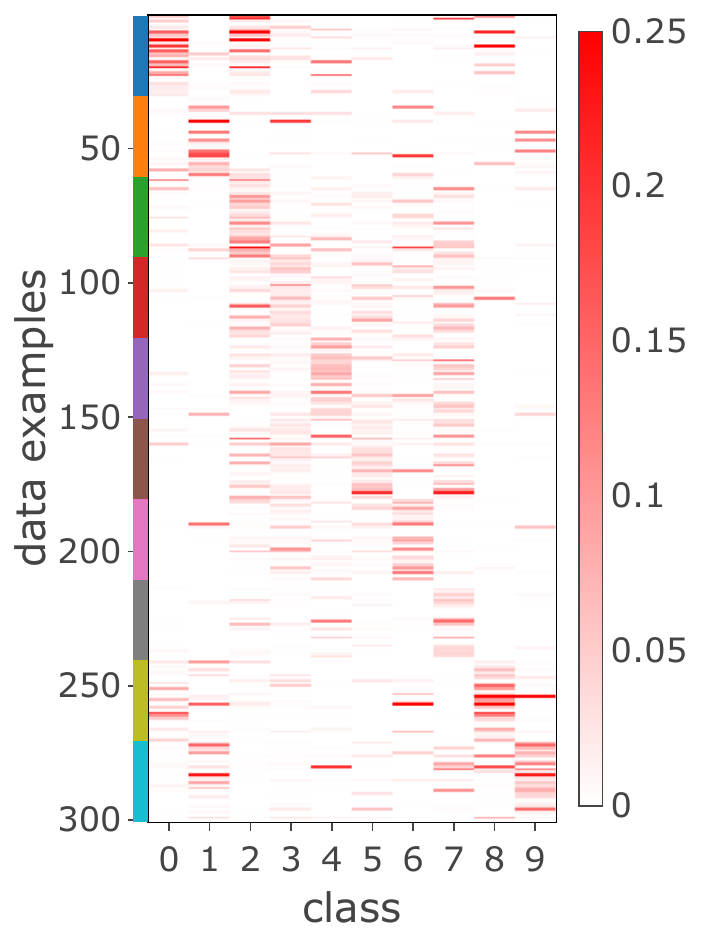}
   \includegraphics[height=1.7in]{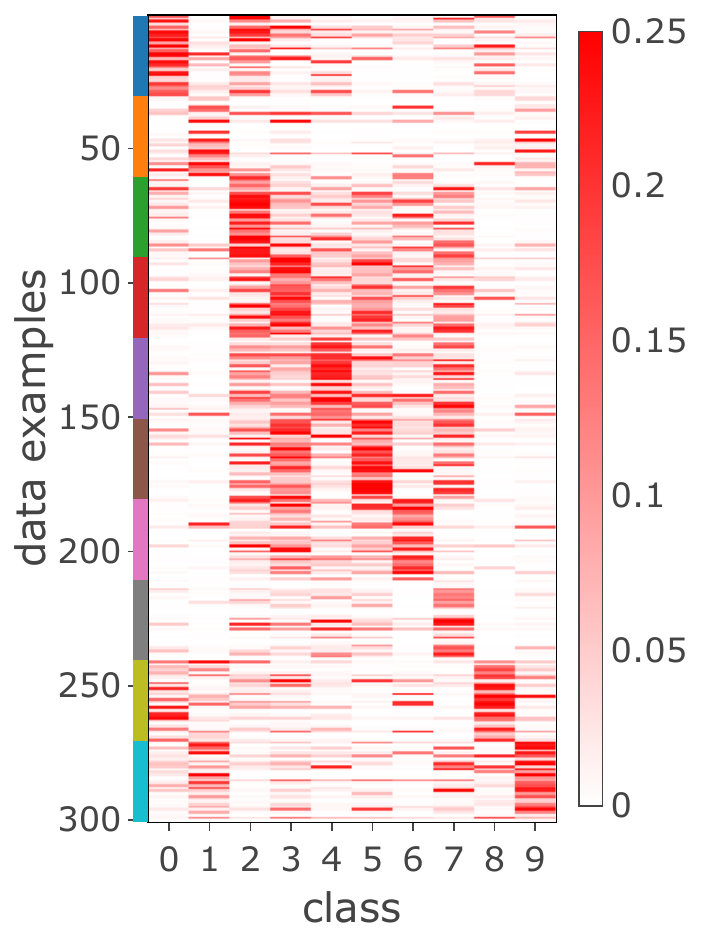}
  \vspace{-3pt}
   \subcaption{Epistemic (left) and aleatoric (right)  uncertainties}
   \label{fig:cifar_unct}
\end{minipage}
\caption{DNN2GP kernels, posterior means and uncertainties with LeNet5 of 300 samples on binary MNIST in Fig. (c), MNIST in Fig. (a), and CIFAR-10 in Fig. (b,d). The colored regions on the y-axis mark the classes.
Fig.~(a) shows the kernel and the predictive mean for the Laplace approximation, which gives 99\% test accuracy. 
We see in the kernel that examples with same class labels are correlated.
Fig.~(c) shows the same for binary MNIST trained only on digits 0 and 1 by using VI.
The kernel clearly shows the out-of-class predictive behavior where predictions are not certain.
Fig.~(b) and (d) show the Laplace-GP on the more complex CIFAR-10 data set where we obtain 68\% accuracy.
Fig.~(d) shows the two components of the predictive variance for CIFAR-10 that can be interpreted as epistemic (left) and aleatoric (right) uncertainties. The estimated epistemic uncertainty is much lower than the aleatoric uncertainty, implying that the model is not flexible enough. This is plausible since the accuracy of the model is not too high (merely 68\%). 
%Fig.~(d) shows how the predictive uncertainty of DNN2GP combines the variance of the function $\vf(\vx)$ and the observations $\vy$. The variance in $\vf(\vx)$ (left) can be low even if the output probabilities of the DNN are uniform, which corresponds to high observation noise $\vepsilon$ (right).
%The noise comes directly from the categorical output probabilities.
%GP kernel and posterior mean corresponding to the Laplace approximation for LeNet5 trained on MNIST and CIFAR-10. The kernel matrix shows the correlations learned by the DNN (classes are grouped and marked with different colors along the axes). For MNIST, a higher posterior mean is assigned to the correct label most of the time (see rows in the middle figure), which reflects the good accuracy obtained by the DNN (99\%). For CIFAR, the accuracy is only 68\%, as a result, the patter is a bit unclear reflecting the uncertainty in the predictions. \newline
%GP posterior mean and kernel corresponding to VI on 2 out of 10 MNIST classes. We clearly see that the \emph{in-class} digits are assigned higher posterior mean and the correlation learned is also significant. There are plenty of other correlations learned by the network due to which we get many overconfident predictions for out-of-training classes, especially on the harder 4 vs. 9 task.
}
\label{fig:kernels_and_predictives}
\vspace{-1em}
\end{figure}

Fig.~\ref{fig:mnist_adam} shows the GP kernel matrix and the posterior mean for the Laplace approximation on MNIST. The rows and columns containing 300 data examples are grouped according to the classes.
The kernel matrix clearly shows the correlations learned by the DNN. As expected, each row in the posterior mean also reflects that the classes are correctly classified (DNN test accuracy is 99\%). 
Fig.~\ref{fig:cifar_adam} shows the GP posterior mean after reparameterization for CIFAR-10 where we see a more noisy pattern due to a lower accuracy of around 68\% on this task.

Fig.~\ref{fig:cifar_unct} shows the two components of the predictive variances that can be interpreted as ``aleatoric'' and ``epistemic'' uncertainty.
As shown in Eq. \eqref{eq:reparam_logistic} in Appendix \ref{app:reparam_logistic}, for a multiclass classification loss, the variance of the prediction of a label at an input $\vx_*$ is equal to $ \vLambda_*(\vx_*) + \vLambda_*(\vx_*)\vJ_*(\vx_*) \widetilde{\vSigma} \vJ_*(\vx_*)^\top\vLambda_*(\vx_*)$.
Similar to the linear basis function model, the two terms here have an interpretation (e.g., see Eq.~3.59 in~\cite{bishop2006pattern}). The first term can be interpreted as the aleatoric uncertainty (label noise), while the second term takes a form that resembles the epistemic uncertainty (model noise). 
Fig.~\ref{fig:cifar_unct} shows these for CIFAR-10 where we see that the uncertainty of the model is low (left) and the label noise rather high (right).
This interpretation implies that the model is unable to flexibly model the data and instead explains it with high label noise.

%In Fig.~\ref{fig:cifar_unct}, we show the decomposition of the corresponding predictive distribution: the variance of a prediction $\vy$ is the sum of the model uncertainty, i.e. the variance of $\vf(\vx)$ (left), and the observation noise, i.e. variance of $\vepsilon$ (right).
%For example, the model can be certain while the categorical uncertainty is high which happens if the model is certain that a sample does not belong to any class.
%68.44
%Due to space constraints, the corresponding results for VI are shown in Appendix~\ref{app:extra_plots}.

In Fig.~\ref{fig:mnistzone}, we study the kernel for classes \emph{outside} of the training dataset using VI. We train LeNet-5 on digits 0 and 1 with VOGN and visualize the predictive mean and kernel on all 10 classes denoted by differently colored regions on the y-axis. We can see that there are slight correlations to the out-of-class samples but no overconfident predictions. In contrast, the pattern between 0 and 1 is quite strong. The kernel obtained with DNN2GP helps to interpret and visualize such correlations.

%We train LeNet-5 using VOGN on two binary-classification problems on MNIST. In both figures, the data examples are sorted according to the digits (0 is on top/left and 9 is at bottom/right). 
%
%Within the MNIST data set, the pair \emph{4} and \emph{9} is one of the hardest to distinguish while pair \emph{0} and \emph{1} forms a simple task. Fig.~\ref{fig:two_digits} shows that the kernel obtained for the simpler task leads to much less correlations with unseen classes and the posterior mean does not produce confident predictions on other classes. However, training on the harder task yields a potentially more complex feature map that leads to overconfident out-of-class predictions and high correlations. These observations are in line with confusion metrics typically obtained on the MNIST data set.\footnote{Exemplary MNIST confusion matrix: \url{https://ml4a.github.io/demos/confusion\_mnist/}}

\subsection{Tuning the Hyperparameters of a DNN Using the GP Marginal Likelihood}
In this section, we demonstrate the tuning of DNN hyperparameters by using the GP marginal likelihood on a real and synthetic regression dataset. 
In the deep-learning literature, this is usually done using cross-validation.
Our goal is to demonstrate that with DNN2GP we can do this by simply computing the marginal likelihood on the \emph{training} set. 

\begin{figure}
\begin{minipage}[b]{.28\textwidth}
  \vspace*{\fill}
         \centering
         \includegraphics[height=1.5in]{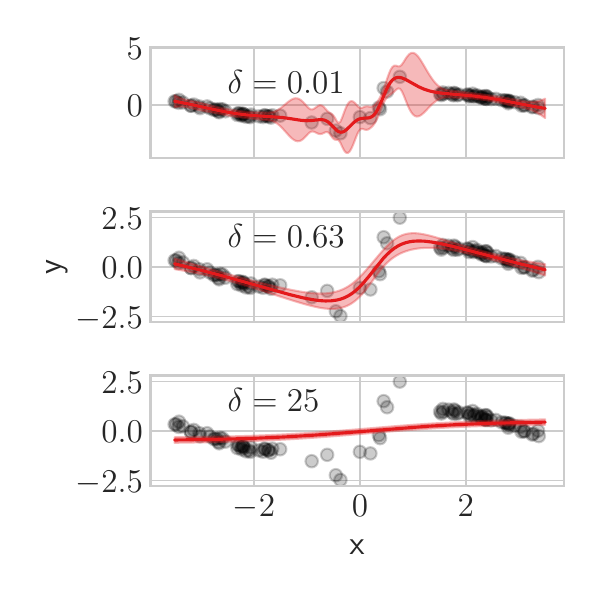}
         \vspace{-10pt}
         \subcaption{Model fits}
         \label{fig:mlh_conv_fits}
\end{minipage}\hfill
\begin{minipage}[b]{.34\textwidth}
  \vspace*{\fill}
         \centering
         \includegraphics[height=1.5in]{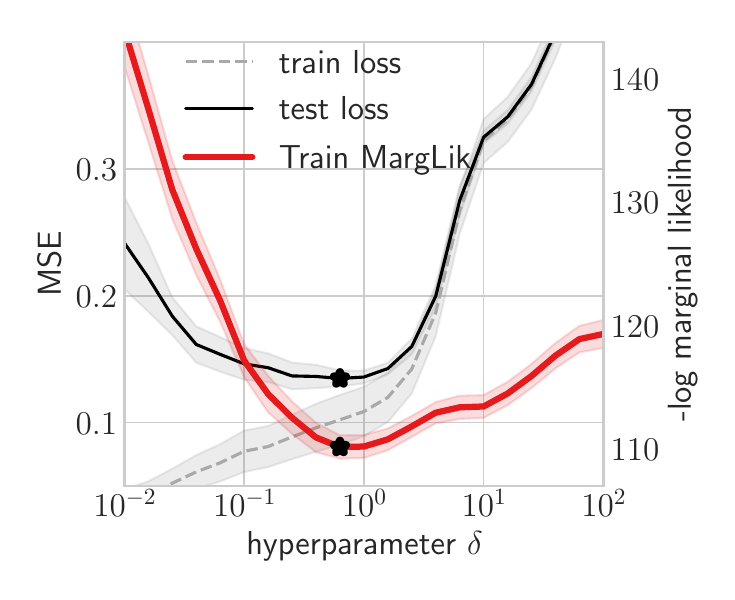}
         \vspace{-10pt}
         \subcaption{Laplace Approximation}
         \label{fig:mlh_conv_lap}
\end{minipage}\hfill
\begin{minipage}[b]{.34\textwidth}
  \vspace*{\fill}
         \centering
         \includegraphics[height=1.5in]{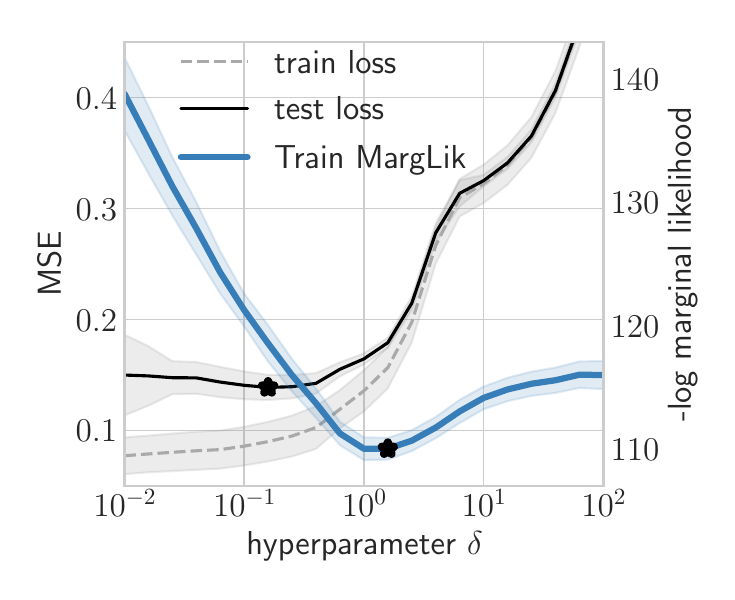}
         \vspace{-10pt}
         \subcaption{Variational Inference}
         \label{fig:mlh_conv_vi}
     \end{minipage}
     \caption{This figure demonstrates the use of the GP marginal likelihood to tune hyperparameters of a DNN. We tune the regularization parameter $\delta$ on a synthetic dataset shown in (a).
     Fig.~(b) and (c) show train and test MSE along with log of the marginal likelihoods on training data obtained with Laplace and VI respectively. We show the standard error over 10 runs. The optimal hyperparameters according to test loss and marginal-likelihood (shown with black stars) match well.}
    \label{fig:mlh}
    \vspace{-1em}
\end{figure}

\begin{figure}
\begin{minipage}[b]{.32\textwidth}
  \vspace*{\fill}
         \centering
         \includegraphics[height=1.43in]{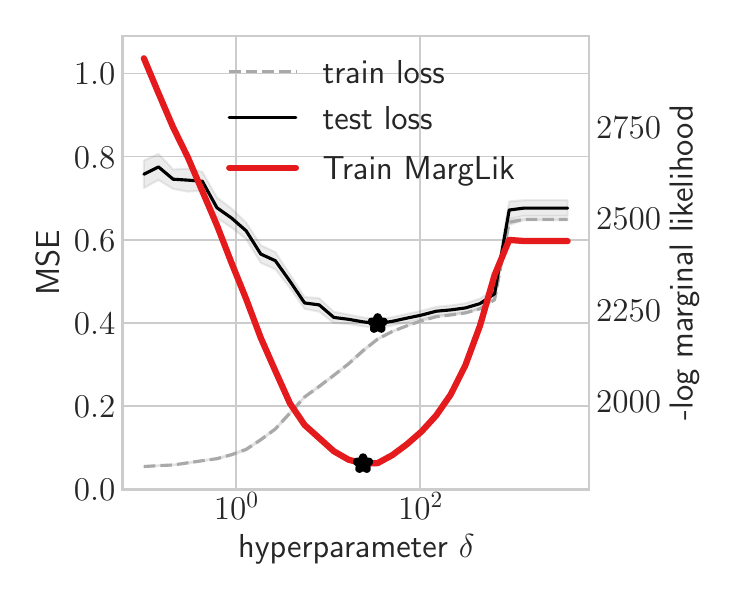}
         \vspace{-20pt}
        %  \subcaption{hyperparameter $\delta$}
        %  \label{fig:wine_delta}
\end{minipage}\hfill
\begin{minipage}[b]{.32\textwidth}
  \vspace*{\fill}
         \centering
         \includegraphics[height=1.43in]{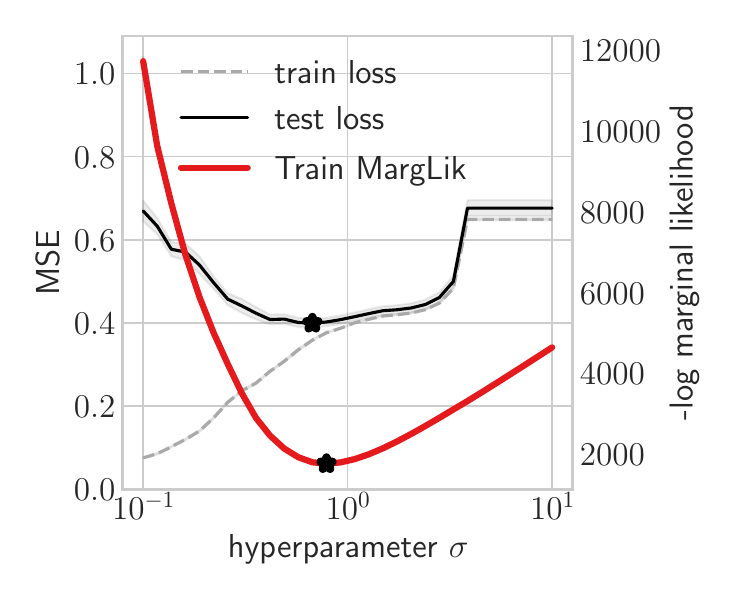}
         \vspace{-20pt}
        %  \subcaption{hyperparameter $\sigma$}
        %  \label{fig:wine_sigma}
\end{minipage}\hfill
\begin{minipage}[b]{.32\textwidth}
  \vspace*{\fill}
         \centering
         \includegraphics[height=1.43in]{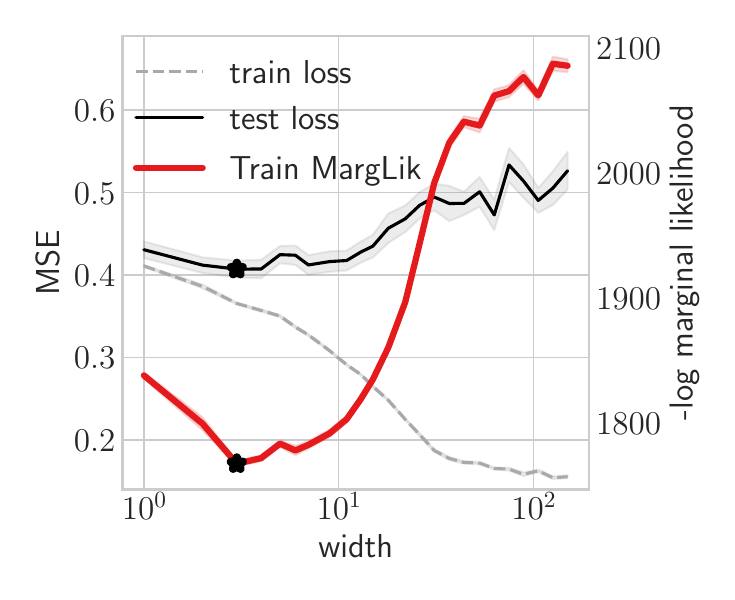}
         \vspace{-20pt}
        %  \subcaption{hidden layer width}
        %  \label{fig:wine_width}
\end{minipage}\hfill
     \caption{This is same as Fig. \ref{fig:mlh} but on a real dataset: UCI Red Wine Quality.
     All the plots use Laplace approximation, and the standard errors are estimated over 20 splits.
     We tune the following hyperparameters: the regularization parameter $\delta$ (left), the noise-variance $\sigma$ (middle), and the DNN width (right).     
     The train log marginal-likelihood chooses hyperparameters that give a low test error.}
    \label{fig:uci_mlh}
    \vspace{-1.5em}
\end{figure}

We generate a synthetic regression dataset ($N=100$; see Fig.~\ref{fig:mlh}) where there are a few data points around $x=0$ but plenty away from it.
We fit the data by using a neural network with single hidden layer of 20 units and $\tanh$ nonlinearity.
Our goal is to tune the regularization parameter $\delta$ to trade-off underfitting vs overfitting.
Fig.~\ref{fig:mlh_conv_lap} and \ref{fig:mlh_conv_vi} show the train log marginal-likelihood obtained with the GP obtained by DNN2GP, along with the test and train mean-square error (MSE) obtained using a point estimate.
Black stars indicate the hyperparameters chosen by using the test loss and log marginal likelihood, respectively.
We clearly see that the train marginal-likelihood chooses hyperparameters that give low test error.
The train MSE on the other hand overfits as $\delta$ is reduced.

Next, we discuss results for a real dataset: UCI Red Wine Quality ($N=1599$) with an input-dimensionality of 12 and a scalar output.
We use an MLP with 2 hidden layers 20 units each and $\tanh$ transfer function.
We consider tuning the regularizer $\delta$, the noise-variance $\sigma$, and the DNN width.
We use the Laplace approximation and tune one parameter at a time while keeping the others fixed (we use respectively $\sigma = 0.64$, $\delta=30$ and $\sigma=0.64$, $\delta=3$, $1$ hidden layer).
Similarly to the synthetic data case, the train marginal-likelihood selects hyperparameters that give low test error.
These experiments show that the DNN2GP framework can be useful to tune DNN hyperparameters, although this needs to be confirmed for larger networks than we used here.

%% file: chapters/_6discussion.tex
In this paper, we present theoretical results connecting approximate inference on DNNs to GP posteriors. Our work enables the extraction  of feature maps and GP kernels by simply training DNNs. It provides a natural way to combine the two different models.

Our hope is that our theoretical results will facilitate further research on combining strengths of DNNs and GPs.
A computational bottleneck is the Jacobian computation which prohibits application to large problems.
There are several ways to reduce this computation, e.g., by choosing a different type of GGN approximation that uses gradients instead of the Jacobians.
Exploration of such methods is a future direction that needs to be pursued.

Exact inference on the GP model we derive is still computationally infeasible for large problems. However, further approximations could enable inference on bigger datasets.
%Our theoretical connections do indicate this to certain degree.
%It is likely that, as the DNN width is increased, approximate inference behaves like the posterior of a GP which is an interesting future direction to pursue.
%
Finally, our work opens many other interesting avenues where a combination of GPs and DNNs can be useful such as model selection, deep reinforcement learning, Bayesian optimization, active learning, interpretation, etc.
We hope that our work enables the community to conduct further research on such problems.

%% file: appendix/_appendix.tex
\section{Proofs}
\label{app:proofs}
\input{appendix/_app_A_proofs.tex}

%\section{Old Proofs (to delete)}
%\label{app:proofs}
%\input{appendix/_app_AA_old_proofs.tex}

\section{Approximating Posterior Predictive with DNN2GP Approach}
\label{app:post_pred}
\input{appendix/_app_B_predictive_dist.tex}

\section{Additional Results}
\label{app:extra_plots}
\input{appendix/_app_C_extra_plots.tex}

\section{Author Contributions Statement}
Author List: Mohammad Emtiyaz Khan, Alexander Immer, Ehsan Abedi, Maciej Korzepa.

M.E.K. conceived a rough idea using the gradients and Hessians of the loss, and wrote the first version of the proofs.
A.I. and E.A. made major corrections to M.E.K.'s original version and introduced version used in the final paper.
They also came up with the prediction method for DNN2GP.
E.A. formalized the NTK connection, and extensively studied its connection to the GP posterior. 
A.I. did most of the experiments and introduced the necessary reparameterization for applications. M.K. helped on the hyperparameter-tuning experiments, as well as with the visualizations.
M.K. did the regression uncertainty experiment with some help from E.A. and A.I.

M.E.K. wrote the main content of the paper. E.A. wrote all the proofs, and A.I. and M.K. summarized the experiment section. All the authors proof-read the paper and revised it.

\section{Camera-Ready Version vs the Submitted Version}
We made several changes taking reviewers feedback into account.

\begin{enumerate}
    \item The writing and organization of the papers were modified to emphasize that we are able to relate the iterations of a deep-learning algorithm to GP inference.
    \item To improve clarity, Fig. 1 was added as a summary of our approach. The writing was modified to follow Step A, B, and C given in Fig. 1.
    \item Titles of Section 3 and 4 were changed to emphasize relationship to "solutions and iterations" of a deep-learning algorithm.
    \item Theorem 1 and 2 were simplified to focus only on the posterior of linear model only. Relation to GP is discussed separately.
    \item Experiment on GP regression was modified to focus on uncertainty instead of the width of the DNN.
    \item Visualization of the GP predictive uncertainty and noise was added on top of predictive mean on CIFAR-10
    \item A real-world experiment on Wine dataset was added, where we tune the width of the DNN.
\end{enumerate}

%% file: appendix/_app_A_proofs.tex
In this section, we prove the theorems presented in the main text.
\subsection{Proof of Theorem \ref{thm:laplace_ggn}} \label{proof:laplaceggn}
We begin with the Gaussian approximation of the Laplace approximation. We will then express its natural parameters in terms of the gradient and Hessians. Application of the GGN approximation and some further manipulation will show that the distribution correspond to the posterior of a linear model.

We start with the Laplace approximation \eqref{eq:laplace_approx} and express its natural parameters in terms of the gradient and Hessians. 
We denote the natural-parameters of this Gaussian approximation $\gauss(\vparam|\vmu,\vSigma)$ by $\veta := \{ \vSigma^{-1}\vmu, -\half\vSigma^{-1} \}$.
In \eqref{eq:laplace_approx}, the second natural parameter is set to the following which is written in terms of the Hessian:
\begin{align}
    -\half \vSigma^{-1} = -\half \sqr{ \sum_{i=1}^N \nabla_{\param\param}^2 \loss_i(\vparam_*) + \delta \vI_P }.
    \label{eq:nat_param_2_laplace}
\end{align}
We can also express the first natural parameter in terms of the gradient and Hessians as shown below. We use the first-order stationary condition, that is, $\nabla_{\param} \bar{\loss}(\data,\vparam_*) = 0$. Adding $\vSigma^{-1}\vmu$ to the both sides of this condition, we get the following:
\begin{align}
    \vSigma^{-1}\vmu &= - \nabla_{\param} \bar{\loss}(\data,\vparam_*) + \vSigma^{-1}\vmu \\
    &= - \sum_{i=1}^N \nabla_\param \loss_i(\vparam_*) -\delta \vparam_*  + \sqr{ \sum_{i=1}^N \nabla_{\param\param}^2 \loss_i(\vparam_*) + \delta \vI_P }\vparam_* \\
    &= \sum_{i=1}^N \sqr{ -\nabla_\param \loss_i(\vparam_*) + \nabla_{\param\param}^2 \loss_i(\vparam_*)  \vparam_* } ,
    \label{eq:nat_param_1_laplace}
\end{align}
where in the second step, we substitute $\vmu$ by $\vparam_*$ and also use \eqref{eq:nat_param_2_laplace}. With this, both natural parameters are now expressed in terms of the gradient and Hessian.

We will now substitute these in the Laplace approximation, denoted by $q_L(\vparam) := \gauss(\vparam|\vmu,\vSigma)$.
Using \eqref{eq:nat_param_2_laplace} and \eqref{eq:nat_param_1_laplace}, we get the following expression: 
\begin{align}
   q_L& (\vparam)   = \frac{1}{\sqrt{(2\pi)^P | \vSigma |}} \exp{\sqr{-\half (\vparam-\vmu)^\top\vSigma^{-1}(\vparam-\vmu)}} \\
   & \propto \exp{ \sqr{  - \half \vparam^\top (\vSigma^{-1})\vparam + \vparam^\top(\vSigma^{-1}\vmu)} } \\
   & = \exp{\rnd{ \frac{-\delta \vparam^\top\vparam}{2} }} \prod_{i=1}^N \exp{ \sqr{ - \half \vparam^\top \nabla_{\param\param}^2 \loss_i(\vparam_*)\vparam + \vparam^\top \crl{ -\nabla_\param \loss_i(\vparam_*) + \nabla_{\param\param}^2 \loss_i(\vparam_*)  \vparam_* }  } },
\end{align}
where in the last line we use \eqref{eq:nat_param_2_laplace} and \eqref{eq:nat_param_1_laplace}.

Now, we will employ the GGN approximation \eqref{eq:ggn} which gives us the Laplace-GGN approximation $\gauss(\vparam|\vmu,\tvSigma)$, shown below:
\begin{equation}
   \exp{\rnd{ \frac{-\delta \vparam^\top\vparam}{2} }} \prod_{i=1}^N \exp{ \sqr{ \frac{-1}{2} \vparam^\top \vJ_*(\vx_i)^\top \vLambda_{i,*} \vJ_*(\vx_i) \vparam +\vparam^\top \vJ_*(\vx_i)^\top\crl{   \vLambda_{i,*} \vJ_*(\vx_i) \vparam_*  - \vr_{i,*} } } },
   \label{eq:tq_L}
\end{equation}
where for notational convenience we have denoted $\vLambda_{i,*} := \vLambda_*(\vx_i,\vy_i)$ and $\vr_{i,*} := \vr_*(\vx_i,\vy_i)$.

A key point here is that each term in the product over $i$ in \eqref{eq:tq_L} is proportional to a Gaussian distribution, provided that $\vLambda_{i,*}\succ 0 $, which is the case since we assume the loss function to be strictly convex.
We will now express each term in the product, as a likelihood over a \emph{pseudo-output} defined as $\tvy_i := \vJ_*(\vx_i)\vparam_* - \vLambda_{i,*}^{-1}\vr_{i,*}$.
Using this and completing the square within each term in the product over $i$ in \eqref{eq:tq_L}, we get the following: 
\be
    \Tilde{q}_L(\vparam) := \gauss(\vparam|\vmu,\tvSigma) \propto \gauss(\vparam|0, \delta^{-1}\vI_P) \prod_{i=1}^N \gauss(\tvy_i | \vJ_*(\vx_i)\vparam, \vLambda_{i,*}^{-1}).
\ee
The right hand side of the above equation is proportional to the posterior distribution $p(\vparam|\widetilde{\data})$ given a transformed dataset $\widetilde{\data} :=\{(\vx_i,\tilde{\vy}_i)\}_{i=1}^N$ of a linear basis-function model
$ \tvy = \vJ_*(\vx)\vparam + \vepsilon$ with Gaussian noise $\vepsilon \sim \gauss(0, (\vLambda_*(\vx,\vy))^{-1})$ and prior distribution $\vparam \sim \gauss(0,\delta^{-1} \vI_P)$.
This completes the proof.

%The predictive distribution of this model can be equivalently written in the function space, giving rise to a GP regression model as follows
%    \begin{align}
%        \tvy = \vf(\vx) + \vepsilon,
%        \quad \textrm{ with } \vf(\vx) \sim \mathcal{GP}\rnd{ 0, \delta^{-1} \vJ_*(\vx) \vJ_*(\vx')^\top} .
%    \end{align}
It is easy to see that the same proof works when using the approximations shown in \eqref{eq:grad_hess_form}. 
In that case, only the steps from \eqref{eq:tq_L} need to be modified.
The proof also holds when a prior other than Gaussian and a model other than DNN is used. \qed

\subsection{Proof of Theorem \ref{thm:oggn}} \label{proof:oggn}
Similarly to the previous section, we start by writing the Gaussian approximation for VON. We will express its natural parameters in terms of the gradient and Hessians. A GGN approximation and some manipulation will show that the distributions found by VON correspond to posteriors of linear models.

The Gaussian approximation at the $t$'th iteration of VON is denoted by $q_t(\vparam) := \gauss(\vparam|\vmu_t,\vSigma_t)$ where $\vSigma_t := (\vS_t + \delta\vI_P)^{-1}$ is obtained from $\vS_t$.
Using this, we can rewrite the updates \eqref{eq:VON_1} and \eqref{eq:VON_2} in terms of $\vmu_{t}$ and $\vSigma_{t}^{-1}$ as follows
\begin{align}
    \vmu_{t+1} &=  \vmu_t - \beta_t \vSigma_{t+1} \sqr{ \sum_{i=1}^N \myexpect_{q_t(\param)} \sqr{ \nabla_\param \loss_i(\vparam) } + \delta \vmu_t }, \label{eq:ngvi_update_1} \\ 
    \vSigma_{t+1}^{-1} &= (1-\beta_t) \vSigma_t^{-1} + \beta_t \sqr{ \sum_{i=1}^N \myexpect_{q_t(\param)} \sqr{ \nabla_{\param\param}^2 \loss_i(\vparam)} + \delta \vI_P  }.
    \label{eq:ngvi_update_2}
\end{align}
It is again more convenient if we can have an update formula for the natural-parameters of the Gaussian distribution $\gauss(\vparam|\vmu_t,\vSigma_t)$, denoted by $\veta_t := \{ \vSigma_t^{-1} \vmu_t, -\half\vSigma_t^{-1} \}$. So we use similar techniques to find an update for $\veta_t$. In addition, since there are no closed-form expressions for the expectations above, we use $S$ number of samples $\vparam_t^{(s)} \sim q_t(\vparam)$, for $s=1,2,\ldots,S$, and use Monte Carlo (MC) approximation. 

 Given \eqref{eq:ngvi_update_2}, the update corresponding to the second natural-parameter is obvious and given by
\begin{align}
    -\half \vSigma_{t+1}^{-1} 
    &= (1-\beta_t) \sqr{-\half \vSigma_t^{-1}} -\half \beta_t \sqr{ \sum_{i=1}^N \myexpect_{q_t(\param)} \sqr{ \nabla_{\param\param}^2 \loss_i(\vparam)} + \delta \vI_P  } \\
    &\approx  (1-\beta_t) \sqr{-\half \vSigma_t^{-1}} -\half \beta_t \sqr{ \frac{1}{S}\sum_{i,s=1}^{N,S} \nabla_{\param\param}^2 \loss_i(\vparam_t^{(s)}) + \delta \vI_P  },
    \label{eq:nat_param_2_vi_iter}
\end{align}
where we have used an MC approximation in the second step.

To write the update for the first natural-parameter, we multiply \eqref{eq:ngvi_update_1} by $\vSigma_{t+1}^{-1}$ and get
\begin{align}
    \vSigma_{t+1}^{-1} \vmu_{t+1}
    &=  \vSigma_{t+1}^{-1} \vmu_t - \beta_t \sqr{ \sum_{i=1}^N \myexpect_{q_t(\param)} \sqr{ \nabla_\param \loss_i(\vparam) } + \delta \vmu_t } \\
    &= (1-\beta_t) \sqr{\vSigma_t^{-1}\vmu_t}
    +  \beta_t \sum_{i=1}^N \sqr{ - \myexpect_{q_t(\param)} \sqr{ \nabla_\param \loss_i(\vparam) }+ \myexpect_{q_t(\param)} \sqr{ \nabla_{\param\param}^2 \loss_i(\vparam)} \vmu_t   } \\
    &\approx (1-\beta_t) \sqr{\vSigma_t^{-1}\vmu_t}
    +  \frac{\beta_t}{S} \sum_{i,s=1}^{N,S} \sqr{ -  \nabla_\param \loss_i(\vparam_t^{(s)}) +  \nabla_{\param\param}^2 \loss_i(\vparam_t^{(s)}) \vmu_t   },
    \label{eq:nat_param_1_vi_iter}
\end{align}
where in the second step, we replaced $\vSigma_{t+1}^{-1}$ in the first term by \eqref{eq:ngvi_update_2}.
The posterior approximation $q_{t+1}(\vparam)$ at time $t+1$ can be written in terms of natural parameters as shown below:
\begin{align}
      q_{t+1}(\vparam) &  = \frac{1}{\sqrt{(2\pi)^P |\vSigma_{t+1}|}} \exp{\sqr{-\half (\vparam-\vmu_{t+1})^\top\vSigma_{t+1}^{-1}(\vparam-\vmu_{t+1})}} \\
      & \propto \exp{ \sqr{  - \half \vparam^\top (\vSigma_{t+1}^{-1})\vparam + \vparam^\top(\vSigma_{t+1}^{-1}\vmu_{t+1}) } } .
\end{align}
By substituting the natural parameters from \eqref{eq:nat_param_2_vi_iter} and \eqref{eq:nat_param_1_vi_iter}, we get the following update for $q_{t+1}(\vparam)$, expressed in terms of the MC samples:
\begin{align}
   q_{t+1}(\vparam) &\propto  p(\vparam)^{\beta_t} q_t(\vparam)^{1-\beta_t} \times \nonumber \\
   &\prod_{i,s = 1}^{N,S} \exp{ \sqr{ - \frac{\beta_t}{2S} \vparam^\top \nabla_{\param\param}^2 \loss_i(\vparam_t^{(s)})\vparam + \frac{\beta_t \vparam_t^\top}{S}  \crl{ -\nabla_\param \loss_i(\vparam_t^{(s)}) + \nabla_{\param\param}^2 \loss_i(\vparam_t^{(s)})  \vmu_t }  } }, \label{eq:product}
\end{align}
where $p(\vparam) = \gauss(\vparam \given 0,\delta^{-1} \vI_P)$ is the prior distribution. For the product of posterior approximation at time $t$ and prior in \eqref{eq:product}, we obtain the following unnormalized Gaussian
\begin{align}
    p(\vparam)^{\beta_t} q_t(\vparam)^{1-\beta_t} 
    =\gauss \rnd{\vparam \given 0,\delta^{-1} \vI_P}^{\beta_t} 
      \gauss \rnd{\vparam \given \vmu_t, \vSigma_t}^{1-\beta_t} \propto\gauss \rnd{\vparam \given \vm_t, \vV_t},
\end{align}
where $\vV_t$ and $\vm_t$ are given by
\begin{equation}
\vV_t^{-1} :=  (1-\beta_t) \vSigma_t^{-1} + \beta_t \delta \vI_P, \quad \vm_t := (1-\beta)\vV_t \vSigma_t^{-1}\vmu_t.    
\end{equation}
Next, for the product over $i$ and $s$ in \eqref{eq:product}, we employ the GGN approximation \eqref{eq:ggn} and get
\begin{align}
    \Tilde{q}_{t+1} & (\vparam)  \propto  \gauss \rnd{\vparam \given \vm_t, \vV_t} \times \nonumber \\
    & \prod_{i,s = 1}^{N,S} \exp{ \sqr{ - \vparam^\top \vJ_{s,t}(\vx_i)^\top \frac{\beta_t\vLambda_{i,s,t}}{2S} \vJ_{s,t}(\vx_i) \vparam +\frac{\beta_t \vparam^\top \vJ_{s,t}(\vx_i)^\top}{S}\crl{   \vLambda_{i,s,t} \vJ_{s,t}(\vx_i) \vmu_t  - \vr_{i,s,t} } } }, \label{eq:tqt_1}
\end{align}
where we have defined $\vJ_{s,t}(\vx_i) := \vJ_{\param_t^{(s)}}(\vx_i)$, $\vr_{i,s,t} := \vr_{\param_t^{(s)}}(\vx_i,\vy_i)$, and $\vLambda_{i,s,t} := \vLambda_{\param_t^{(s)}}(\vx_i,\vy_i)$. The notation $\Tilde{q}_{t+1}(\vparam)$ is used to emphasize that GGN approximation is used in this update.

We are now ready to express each term in the product above as a Gaussian distribution.
First, we define three quantities: $\vJ_t(\vx), \vr_t(\vx,\vy)$ and $\vLambda_t(\vx,\vy)$ which are obtained by concatenating all the sampled Jacobians, residuals, and noise-precision matrices:
\begin{align}
    \vJ_t(\vx) &:=
    \sqr{
    \begin{array}{c}
         \vJ_{\param_t^{(1)}}(\vx) \\
         \vJ_{\param_t^{(2)}}(\vx) \\
         \vJ_{\param_t^{(3)}}(\vx) \\
         \vdots\\
         \vJ_{\param_t^{(S)}}(\vx) \\
    \end{array}},
    \quad\quad 
    \vr_t(\vx,\vy) :=
    \sqr{
    \begin{array}{c}
         \vr_{\param_t^{(1)}}(\vx,\vy) \\
         \vr_{\param_t^{(2)}}(\vx,\vy) \\
         \vr_{\param_t^{(3)}}(\vx,\vy) \\
         \vdots\\
         \vr_{\param_t^{(S)}}(\vx,\vy) \\
    \end{array}}, \\
    \vLambda_t(\vx,\vy) &:=
    \sqr{
    \begin{array}{ccccc}
         \vLambda_{\param_t^{(1)}}(\vx,\vy) & 0 & 0 & \ldots & 0 \\
         0 & \vLambda_{\param_t^{(2)}}(\vx,\vy) & 0 & \ldots & 0 \\
         0 & 0 & \vLambda_{\param_t^{(3)}}(\vx,\vy) & \ldots & 0 \\
         \vdots & \vdots & \ddots & \vdots & \vdots \\
         0 & 0 & 0 & \ldots & \vLambda_{\param_t^{(S)}}(\vx,\vy)\\
    \end{array}}.
\end{align}
Using this, we define a transformed output of length $KS \times 1$ as 
\begin{align}
    \tvy_{i,t} := \vJ_{t}(\vx_i)\vmu_t - \vLambda_t(\vx_i,\vy_i)^{-1}\vr_t(\vx_i,\vy_i).
\end{align}
The distribution $\Tilde{q}_{t+1}(\vparam)$ defined in \eqref{eq:tqt_1} can then be expressed as the following:
\begin{equation}
    \Tilde{q}_{t+1} (\vparam) \propto \gauss \rnd{\vparam \given \vm_t, \vV_t} \prod_{i=1}^{N} \gauss\rnd{\tvy_{i,t} | \vJ_t(\vx_i)\vparam, S (\beta_t\vLambda_t(\vx_i,\vy_i))^{-1}}.
\end{equation}
As before, we can show that this distribution is equal to the posterior distribution of a linear on a transformed dataset defined as $\widetilde{\data}_{t} :=\{(\vx_i,\tilde{\vy}_{i,t})\}_{i=1}^{N}$.
To model such outputs, we define a linear model for an output $\tvy_t \in \real^{KS}$ defined as follows: 
\begin{align}
\tvy_t = \vJ_t(\vx)\vparam + \vepsilon_{t}, \textrm{ with } \vepsilon_{t} \sim \gauss(0, S (\beta_t\vLambda_t(\vx,\vy))^{-1}), \textrm{ and } \vparam \sim \gauss \rnd{\vm_t, \vV_t}.
\end{align} 
The theorem presented in the main text is a simpler version of this theorem where $S = 1$. This completes the proof.
\qed

\subsection{Linear Model Corresponding to OGGN}
\label{app:oggn_proof}
In OGGN, we evaluate the gradient and Hessian at the mean $\vmu_t$ defined to be equal to the current iterate $\vparam_t$.
This corresponds to $S=1$ in the setting described in the proof of theorem \ref{thm:oggn} (see Appendix \ref{proof:oggn}) with $\vparam_t^{(1)} := \vparam_t$. Therefore, the linear model is the same as before but with $\vJ_t(\vx), \vr_t(\vx,\vy)$ and $\vLambda_t(\vx,\vy)$ defined at $\vparam_t$.

%The predictive distribution of this linear model is equivalent to that of the following GP regression model:
%\begin{align}
%        \tvy_{s} = \vf_s(\vx) + \vepsilon_{s},
%        \quad \textrm{ with } \vf_s(\vx) \sim \mathcal{GP}\rnd{ \vJ_s(\vx)\vm_t, \vJ_s(\vx) \vV_t \vJ_{s'}(\vx')^\top} .
%\end{align}
%\todo[size=\tiny]{here}
%Finally, if, in particular, we set $S=1$ and employ zeroth-order delta approximation, the updates \eqref{eq:VON_1} and \eqref{eq:VON_2} turn to the updates \eqref{eq:ON_2}.
%So derivations in this subsection hold true and we recover the linear model \eqref{eq:lin_model_oggn}. This completes the proof. 

%% file: appendix/_app_B_predictive_dist.tex
Typically, we can always predict using Monte Carlo sampling from the Gaussian approximation, however, this might be too noisy sometimes. 
In this section, we show how DNN2GP approach enables us to directly use the GP regression model for \emph{approximating} the posterior predictive distribution. We elaborate on the method for Laplace approximation but this can be generalized to VI as briefly explained in \autoref{predictive_VI}. 

Given a test input, denoted by $\vx_*$, we first compute the feature map $\vJ_{*}(\vx_*)^\top$. Using the linear model found in the DNN2GP approach, we can compute the posterior predictive distribution of the output, which we denote by $\tvy_*$.
However, to be able to compute the predictive distribution for the true output $\vy_*$, we need to \emph{invert the map} from $\vy_*$ to $\tvy_*$.
The expressions for this map can be obtained by using the definition of the transformed output $\tvy_* := \vJ_*(\vx_*) \vparam_* - \vLambda_*(\vx_*,\vy_*)^{-1}\vr_*(\vx_*,\vy_*)$. We demonstrate this for two common cases of squared loss and logistic loss.

\subsection{Laplace Approximation and Squared Loss}
Consider the squared loss, $\loss(\vy,\vf_\param(\vx)) = \frac{1}{2\sigma^2} \| \vy - \vf_\param(\vx) \|^2$ with $\sigma^2$ as the noise variance. According to \autoref{sec:laplace_dlgp}, in this case, we have $\vr_*(\vx,\vy) := \sigma^{-2} (\vf_{\param_*}(\vx) - \vy)$ and $\vLambda_*(\vx,\vy) := \sigma^{-2} \vI_K$.
Using these expressions in the definition for $\tvy := \vJ_*(\vx)\vparam_* - \vLambda_*(\vx,\vy)^{-1} \vr_*(\vx,\vy)$, we get the following map for the test input $\vx_*$:
\begin{align}
    \tvy_* &= \vJ_*(\vx_*) \vparam_* - (\vf_{\param_*}(\vx_*) - \vy_*)\\
    \implies \vy_* &= \tvy_* + \vf_{\param_*}(\vx_*) - \vJ_{*}(\vx_*)\vparam_*
    \label{eq:GPmodel_squared_loss}
\end{align}

Given a predictive distribution for $\tvy_*$ computed by the linear model \eqref{eq:lin_laplace_ggn} with the posterior distribution $\gauss(\vparam|\vparam_*,\widetilde{\vSigma})$, we can therefore derive the predictive distribution for $\vy_*$. 
In the example above, the predictive variance of $\tvy_*$ and $\vy_*$ will be the same, while the predictive mean of $\vy_*$ is obtained by adding $\vf_{\param_*}(\vx_*) - \vJ_{*}(\vx_*)\vparam_*$ to the mean of $\tvy_*$. The result is as follows
\begin{align}
    \vy_* |\vx_*, \data \sim \gauss\rnd{\vy_* |  \vf_{\param_*}(\vx_*),   \vJ_*(\vx_*) \widetilde{\vSigma} \vJ_*(\vx_*)^\top + \sigma^{2} \vI_K} .
\end{align}
We use this technique to compute the predictive distribution in Fig.~\ref{fig:Snelson} (labeled as `DNN2GP' in the plots).

%Using the expression above in \eqref{eq:GPmodel_squared_loss}, we can compute the predictive mean and variance of $\vy_{test}$ as it is displayed in Fig.~\ref{fig:Williams_infinite_width}.

\subsection{Laplace Approximation and Logistic Loss}\label{app:reparam_logistic}
The procedure above for \emph{inversion of maps} generalizes to other loss functions derived using generalized linear models.
We need to assume that the loss corresponds to a log probability distribution, i.e., $ \loss(\vy, \vf_\param(\vx)) := -\log p(\vy \given \vh(\vf_\param(\vx)) )$ where $\vh(\cdot)$ is a \emph{link function}.
We now describe this for a Bernoulli distribution $y_i \in \{0,1\}$ using the results in \autoref{sec:laplace_dlgp}.

Similarly to the squared-loss case, we need to write $\ty$ in terms of the true output $y$.  
For a Bernoulli likelihood, the link function is $\sigma(f_{\param_*}(\vx)) =: p_*(\vx) $ where $\sigma$ is the sigmoid function, the residual is $r_*(\vx,y) = p_*(\vx) - y$, and the noise precision is $\Lambda_{\param_*}(\vx,y) = p_*(\vx) (1 - p_*(\vx)) := \lambda_*(\vx)$. We again use the definition for the transformed output and write the map for the test input $\vx_*$: 
\begin{align}
    \ty_* &= \vJ_*(\vx_*)\vparam_* - \lambda_*(\vx_*)^{-1} (p_*(\vx_*)-y_*)  \\
    \implies y_* &= p_*(\vx_*) + \lambda_*(\vx_*) \ty_* - \lambda_*(\vx_*) \vJ_*(\vx_*) \vw_*  \label{eq:invbern}
\end{align}
Given the predictive distribution over $\ty_*$ at the test input $\vx_*$, we can then compute the corresponding distribution over $y_*$.
The predictive distribution of $\tilde{y}_*$ in the linear model \eqref{eq:lin_laplace_ggn} with the posterior distribution $\gauss(\vparam|\vparam_*,\widetilde{\vSigma})$ is given as follows: 
\begin{align}
    \ty_* | \vx_*, \widetilde{\data} \sim \gauss \rnd{\ty_*| \vJ_*(\vx_*)\vparam_*, \lambda_*(\vx_*)^{-1} + \vJ_*(\vx_*) \widetilde{\vSigma} \vJ_*(\vx_*)^\top  } .
\end{align}
Therefore, using the map \eqref{eq:invbern}, we get the following predictive distribution over $y_*$:
\begin{align}\label{eq:reparam_logistic}
    y_* |\vx_*, \data \sim \gauss\rnd{y_* |  \sigmoid(f_{\param_*}(\vx_*)),  \lambda_*(\vx_*) + \lambda_*(\vx_*)^2\vJ_*(\vx_*) \widetilde{\vSigma} \vJ_*(\vx_*)^\top } .
\end{align}
Similar to the linear basis function model, the two terms in the predictive variance have an interpretation (e.g., see \cite{bishop2006pattern} Eq. 3.59).
The first term can be interpreted as the aleatoric uncertainty (label noise), while the second term takes a form that resembles the epistemic uncertainty (model noise). 
Such interpretation is possible due to the conversion of a DNN to a linear-bassis function model in our DNN2GP framework.
%Fig.~\ref{fig:cifar_unct} shows these for CIFAR-10 where we see that the uncertainty of the model is low (left) and the label noise rather high (right).
%This interpretation then implies that the model is unable to flexibly model the data, thereby giving a high label noise.

This approach can be similarly written for other Gaussian approximations.
It can also be generalized to loss functions obtained using the generalized linear model. The inversion of the map is possible whenever the link function $\vh(.)$ is invertible.

\subsection{Generalization to VI}\label{predictive_VI}
For the VOGGN update with one MC sample, we can use the same procedure as above. The same is true for OGGN since one MC sample is replaced by the mean. For VOGGN with multiple MC samples, we get $S$ such maps. Each of those maps give us a prediction, denote it by $\tvy_{*,s,t}$ for sample $s$ at iteration $t$. To obtain the final prediction, we can use the average all predictions $\tvy_{*,s,t}$ over $s=1,2,3,\ldots,S$ to get the predictive distribution for $\vy_{*,t}$.

%% file: appendix/_app_C_extra_plots.tex
In this appendix, we provide additional figures to the ones presented in Sec.~\ref{sec:appliedexp}. 

\subsection{Further Posteriors and Kernels for MNIST and CIFAR}
\begin{figure}[!htb]
\begin{minipage}[b]{.66\textwidth}
  \vspace*{\fill}
  \centering
  \includegraphics[height=1.7in]{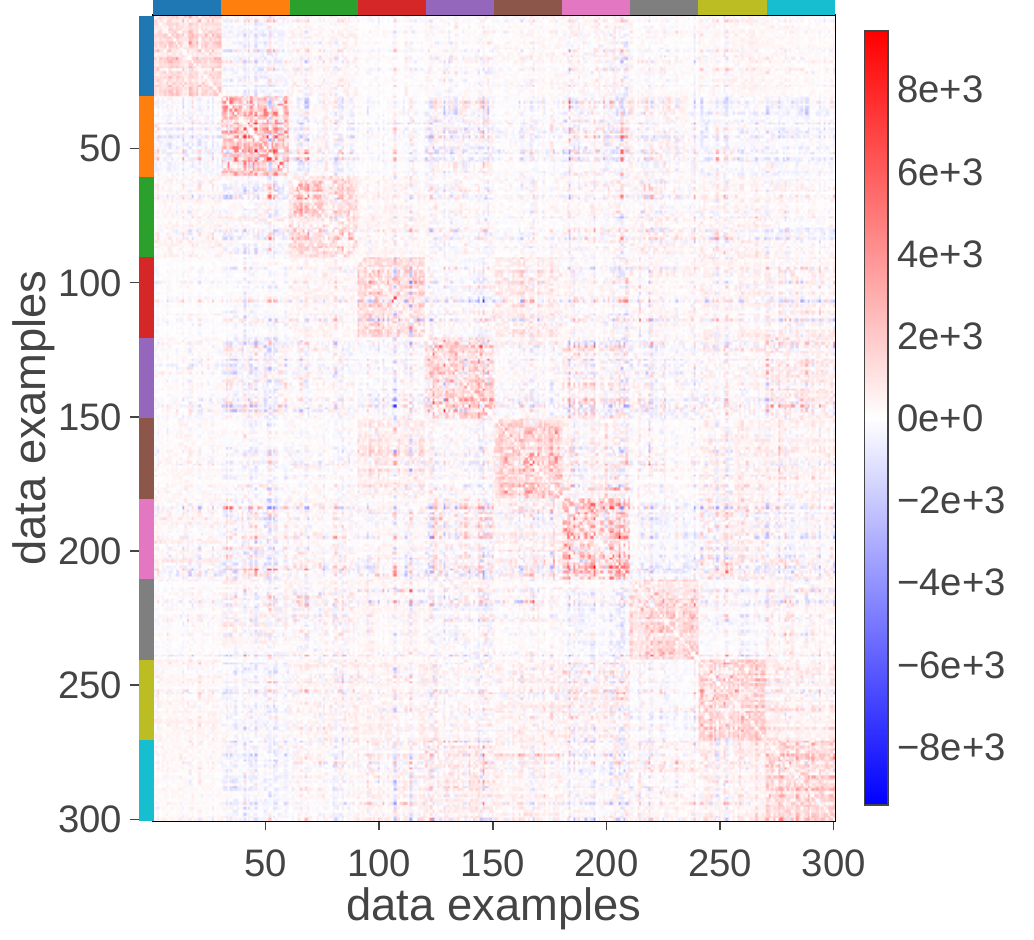}
    \includegraphics[height=1.7in]{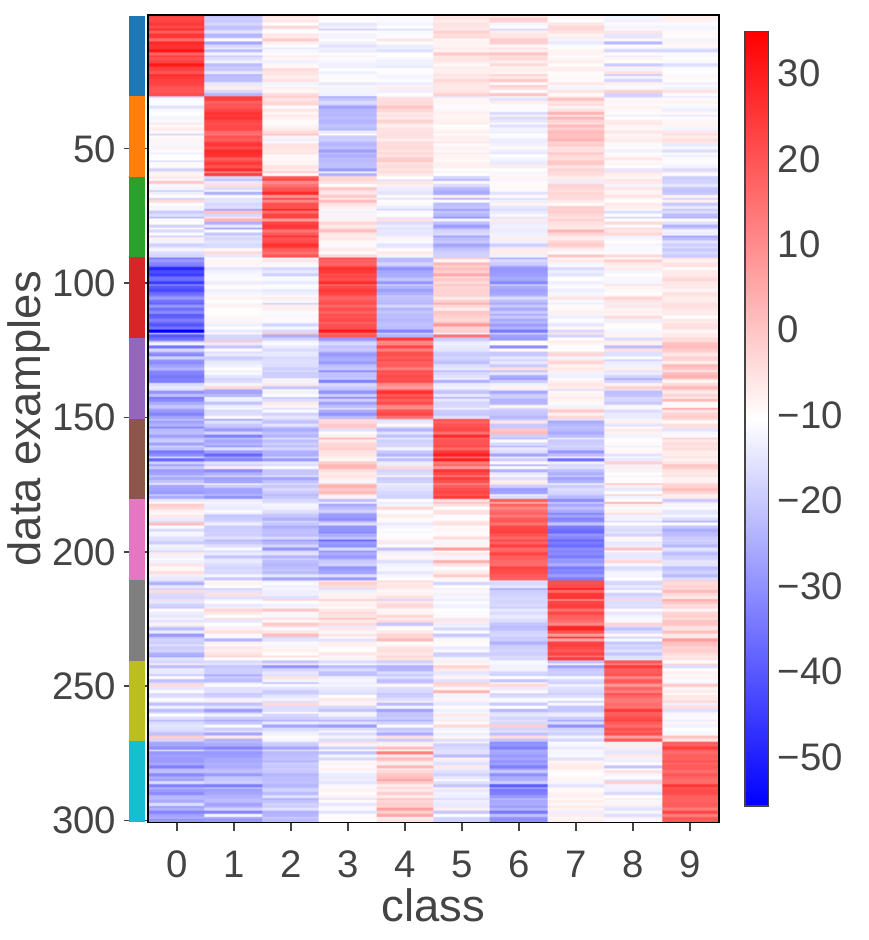}
  \subcaption{MNIST: GP kernel matrix (left) and GP posterior mean (right)}
\end{minipage}\hfill
\begin{minipage}[b]{.32\textwidth}
  \vspace*{\fill}
  \centering
\includegraphics[height=1.7in]{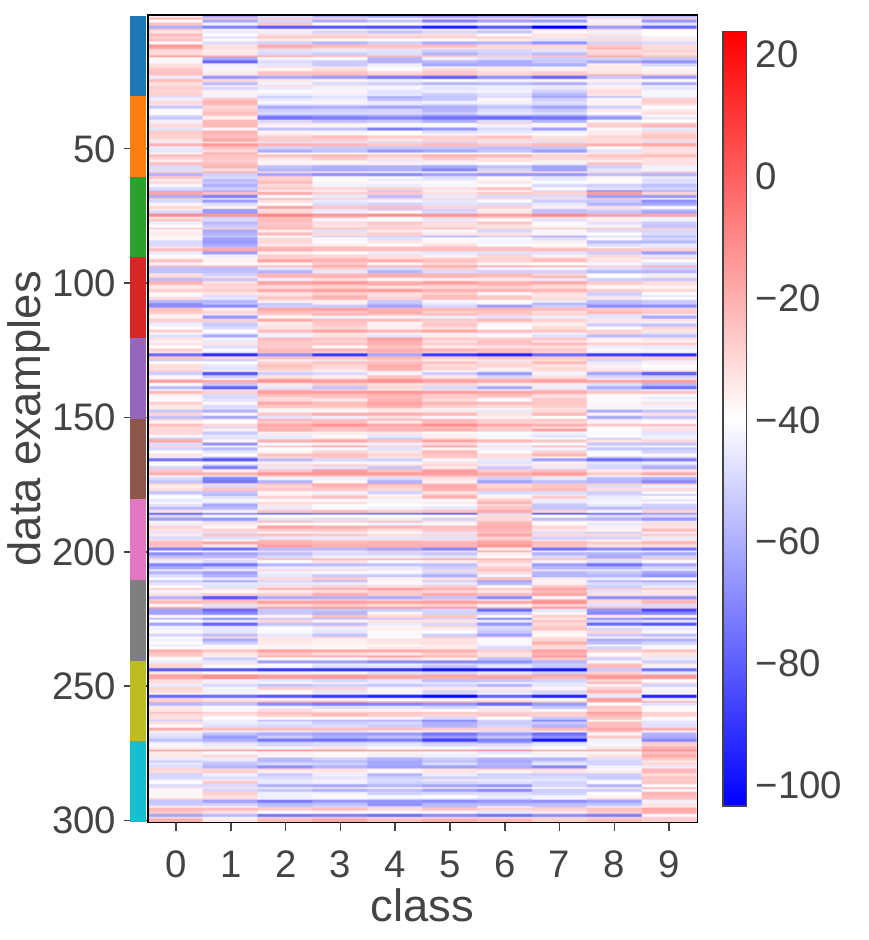}
  \subcaption{CIFAR-10: GP posterior mean}
\end{minipage}
\caption{This figure visualizes the GP kernel matrix and posterior mean for LeNet5 trained with VOGN on MNIST (left) and CIFAR-10 (right). The kernel matrix clearly shows the correlations learned by the DNN. A higher posterior mean is assigned to the correct label which reflects the accuracy obtained by the DNN.}
\label{fig:vogn_mnist_cifar}
\end{figure}

\begin{figure}[!htb]
     \centering
     \begin{subfigure}[b]{0.4\textwidth}
         \centering
  \includegraphics[height=1.7in]{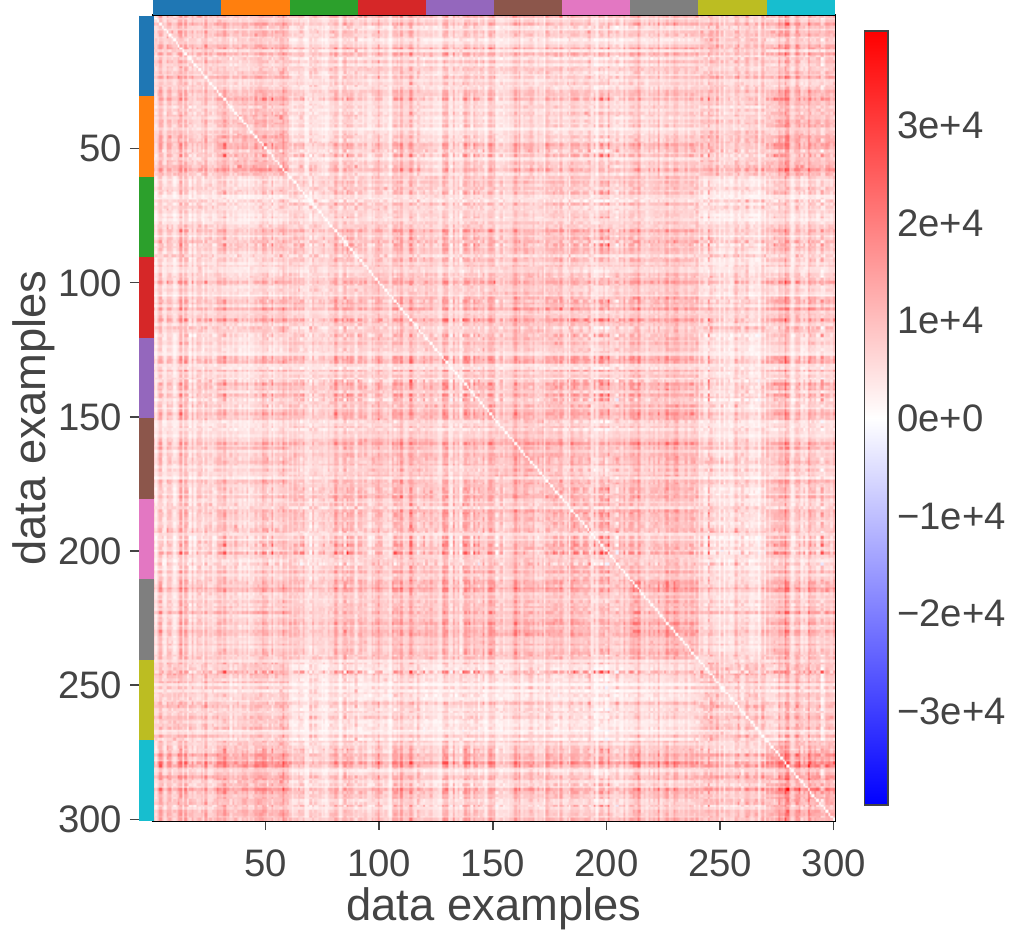}
  \caption{Laplace Approximation}
     \end{subfigure}
     \begin{subfigure}[b]{0.4\textwidth}
         \centering
  \includegraphics[height=1.7in]{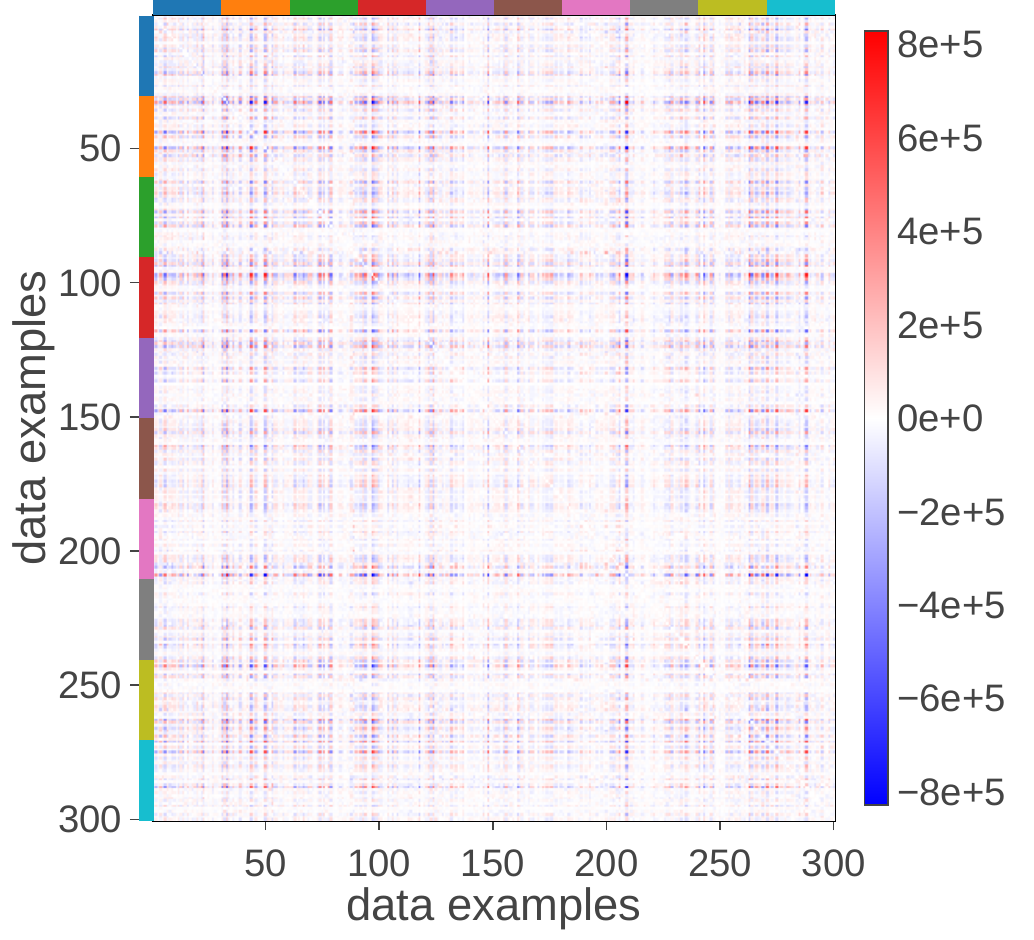}
  \caption{Variational Approximation}
     \end{subfigure}
        \caption{GP kernels due to Laplace and variational approximation for neural networks on CIFAR-10. The kernels show slight traces of structure but are not as significant as the ones presented on MNIST in Sec.~\ref{sec:experiments}.}
        \label{fig:cifar_gp_kernel}
\end{figure}

Fig.~\ref{fig:vogn_mnist_cifar} is similar to Fig.~\ref{fig:mnist_adam} but uses the variational approximation instead of a Laplace approximation. While the posterior mean on MNIST shows very similar structure for both approximations, the kernel shows some interesting differences. There are many more negative correlations between examples from different classes in the kernel corresponding to the variational approximation. The posterior mean on CIFAR-10 has similar structure yet it appears to exhibit higher uncertainty. In Fig.~\ref{fig:cifar_gp_kernel}, we show the kernel matrix on 300 data points of CIFAR-10 with the respective class labels. The kernel is computed for both the Laplace and variational approximation but shows less structure than that of the MNIST dataset.

\subsection{Uncertainties according to DNN2GP for Classification}
\begin{figure}
\begin{minipage}[b]{.32\textwidth}
  \vspace*{\fill}
         \centering
         \includegraphics[height=1.5in]{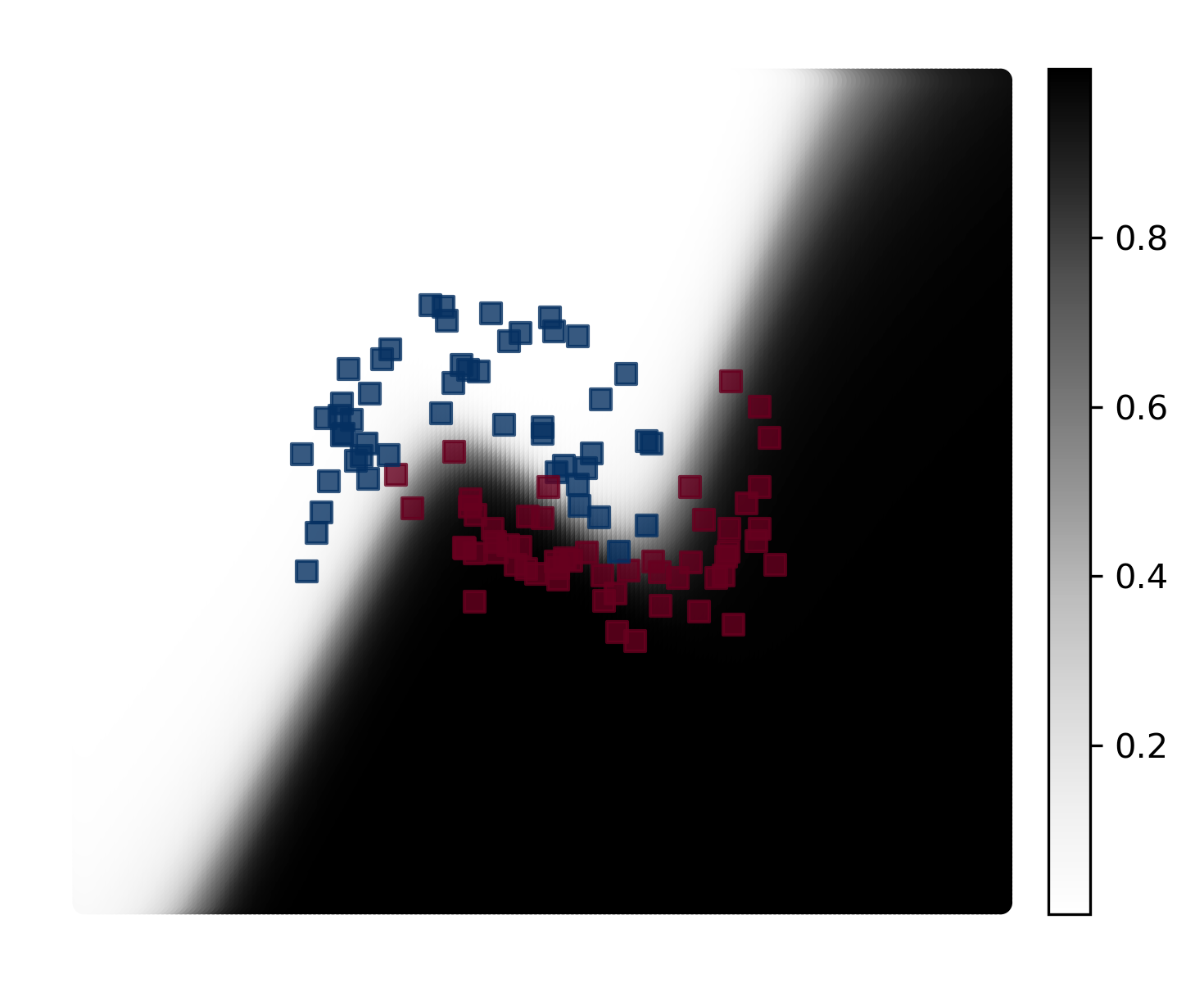}
         \vspace{-10pt}
         \subcaption{mean $\sigmoid(f_{\param_*}(\vx_*))$}
         \label{fig:bin_uct_pred}
\end{minipage}\hfill
\begin{minipage}[b]{.32\textwidth}
  \vspace*{\fill}
         \centering
         \includegraphics[height=1.5in]{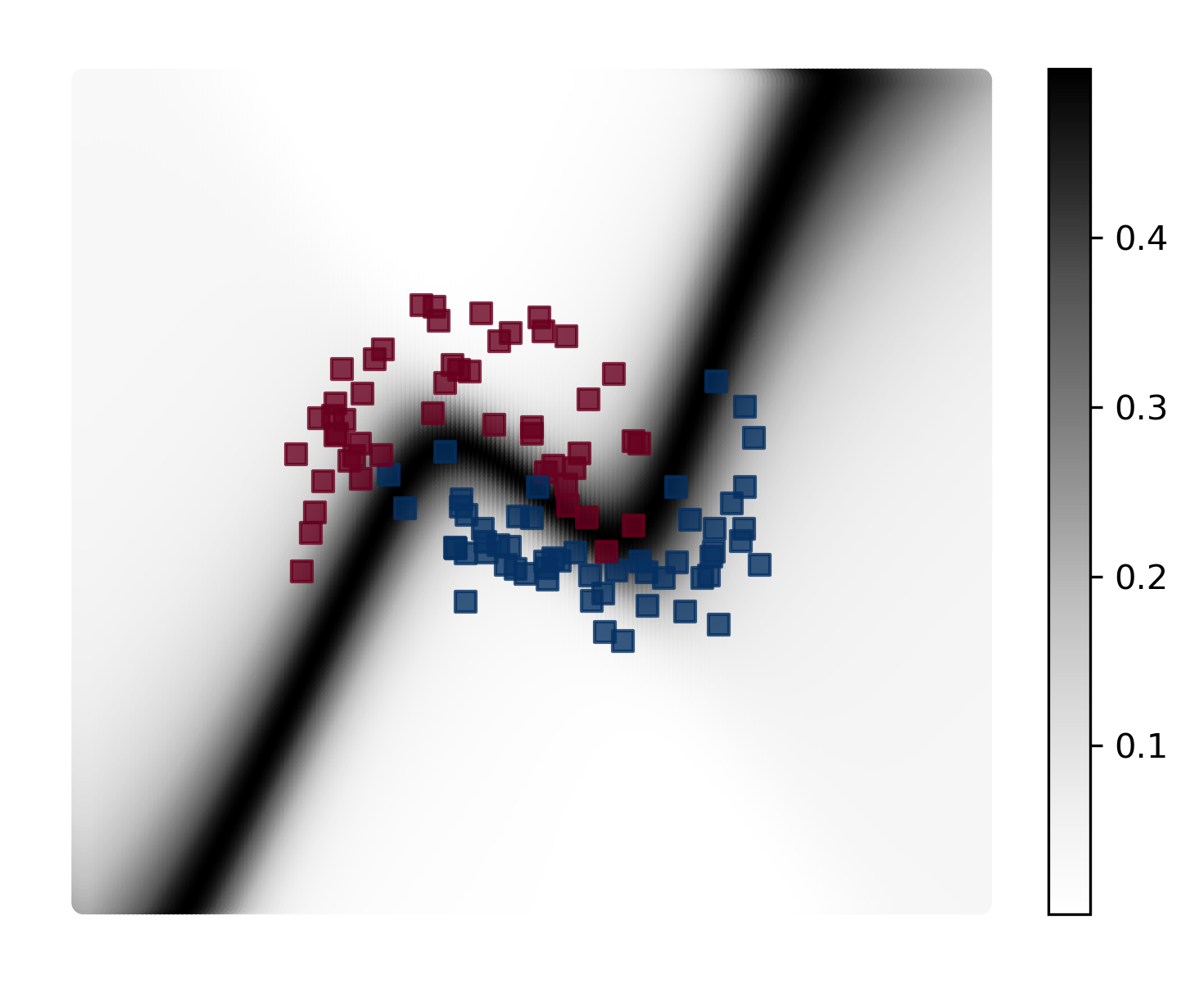}
         \vspace{-10pt}
         \subcaption{aleatoric uncertainty}
         \label{fig:bin_uct_lam}
\end{minipage}\hfill
\begin{minipage}[b]{.32\textwidth}
  \vspace*{\fill}
         \centering
         \includegraphics[height=1.5in]{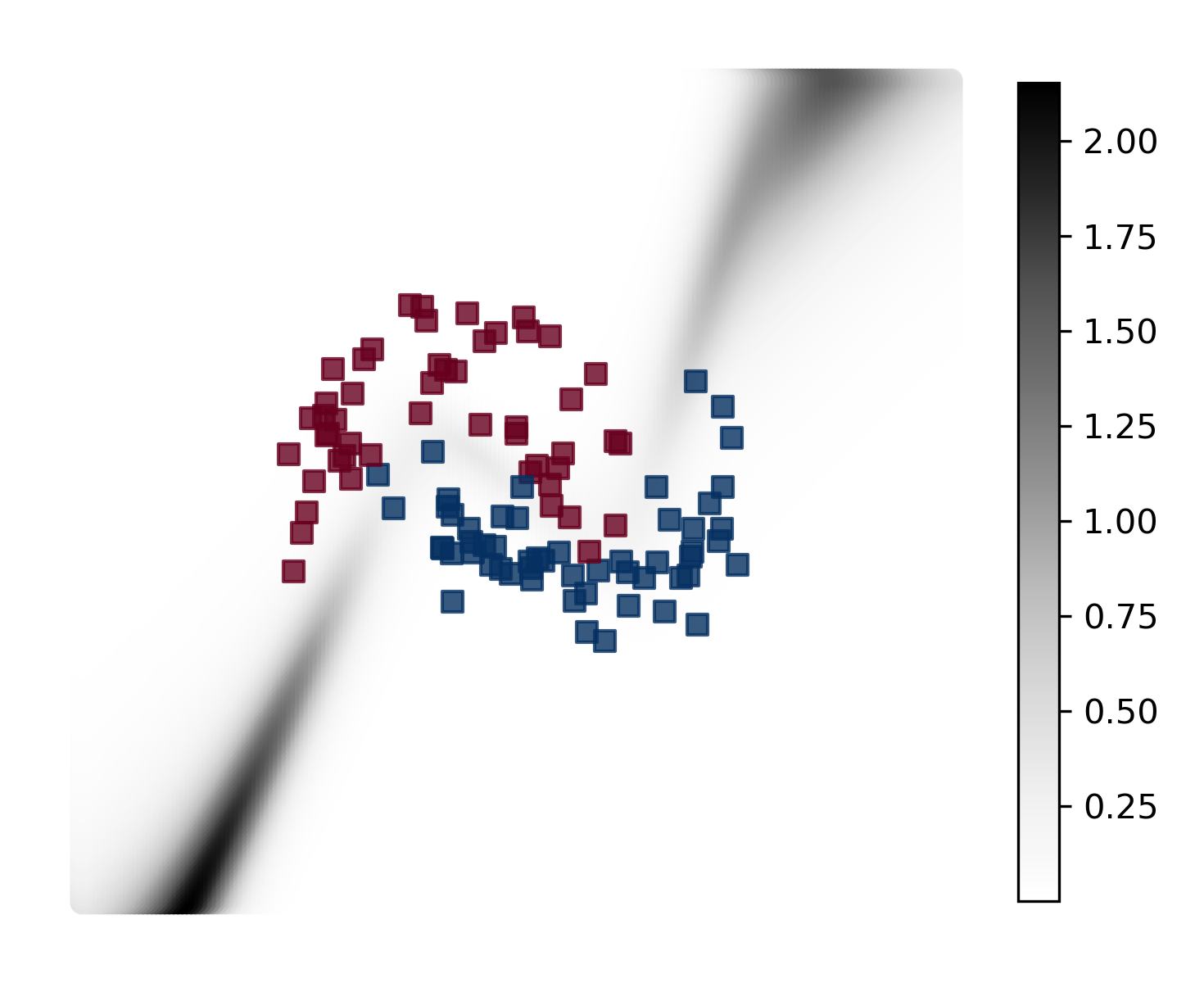}
         \vspace{-10pt}
         \subcaption{epistemic uncertainty}
         \label{fig:bin_uct_jsj}
     \end{minipage}
     \caption{This figure demonstrates the decomposition of predictive variances due to the reparameterization introduced in App.~\ref{app:reparam_logistic} on a binary toy classification task (red vs. blue half moons). We plot the quantities of Eq.~\eqref{eq:reparam_logistic} in figures (a)-(c): (a) is the prediction of a trained NN while the sum of (b) and (c) give us the posterior predictive uncertainties. Around the decision boundary, the label noise~(b) is high and remains unchanged further from the data while the predictive uncertainty is low where supported by data and strongly grows away from it. Here, the model fits the data well in contrast to Fig.~\ref{fig:cifar_unct} where the model is unable to do so which results in high estimated label noise.}
    \label{fig:bin_uct}
    \vspace{-1em}
\end{figure}

In this section we present a toy example for the classification task in line with the regression experiment in Fig.~\ref{fig:Snelson}. We use the reparameterization introduced in App.~\ref{app:reparam_logistic}, in particular Eq.~\eqref{eq:reparam_logistic}. We train a neural network with single hidden layer of 10 units and tanh activation to fit the non-linear decision boundary. We have $\delta=0.26$ and train on 100 samples for 5000 full-batch epochs. Fig.~\ref{fig:bin_uct} shows how the reparameterization allows to decompose predictive variance into label noise due to the decision boundary, see. (b), and model uncertainty, see (c), that grows away from the data.

